\documentclass{article} % For LaTeX2e
\usepackage{iclr2023_conference,times}

\usepackage{url}

\usepackage{amsmath,amsthm,amsfonts,thmtools}
\usepackage{amssymb}
\usepackage{natbib}
\usepackage{graphicx}
\usepackage{listings}
\usepackage{euscript}
\usepackage{units}
\usepackage{enumerate}
\usepackage{wrapfig}
\usepackage{xcolor}
\usepackage{todonotes}
\usepackage{booktabs}
\usepackage{comment}
\usepackage{tikz}

\usepackage{multirow}

\usepackage{mathtools}
\usepackage{subcaption} 
\usepackage{graphicx}
\usepackage{pgfplots}
\usepackage{float}
\usepackage{algpseudocode}
\usepackage{microtype}
\usepackage{rotating}
\usepackage{enumitem}
\usepackage{bm}
\usepackage{hyperref}

\newcommand*{\ldblbrace}{\{\mskip-5mu\{}
\newcommand*{\rdblbrace}{\}\mskip-5mu\}}

\newcommand{\by}{{\bf y}}

\newcommand{\bx}{{\bf x}}
\newcommand{\bX}{{\bf X}}

\newcommand{\bA}{{\bf A}}

\newcommand{\bY}{{\bf y}}
\newcommand{\bF}{{\bf F}}

\newcommand{\bc}{{\bf c}}

\newcommand{\cG}{\mathcal{G}}
\newcommand{\cN}{\mathcal{N}}
\newcommand{\cV}{\mathcal{V}}
\newcommand{\cE}{\EuScript{E}}

\newcommand{\R}{{\mathbb R}}

\newcommand{\fref}[1] {Fig.~\ref{#1}}
\newcommand{\Tref}[1]{Table~\ref{#1}}

\newcommand{\obtau}{\overline{\boldsymbol{\tau}}}

\newcommand{\btau}{\boldsymbol{\tau}}
\newcommand{\cP}{\EuScript{P}}

\newcommand{\model}{{\bf G$^2$}}

\newtheorem{theorem}{Theorem}[section]

\newtheorem{proposition}[theorem]{Proposition}

\newtheorem{remark}[theorem]{Remark}

\title{Gradient Gating for Deep Multi-Rate \\Learning on Graphs}

\author{T. Konstantin Rusch \\
ETH Z\"urich, ICSI and UC Berkeley\\
\texttt{konstantin.rusch@sam.math.ethz.ch}\\
\And
Benjamin P. Chamberlain \\
Charm Therapeutics\\
\And
Michael W. Mahoney \\
ICSI, LBNL, and UC Berkeley\\
\And
Michael M. Bronstein  \\
University of Oxford\\
\And 
Siddhartha Mishra\\
ETH Z\"urich
}

\iclrfinalcopy % Uncomment for camera-ready version, but NOT for submission.
\begin{document}

\maketitle

\begin{abstract}
We present Gradient Gating (\model), a novel framework for improving the performance of Graph Neural Networks (GNNs). Our framework is based on gating the output of GNN layers with a mechanism for multi-rate flow of message passing information across nodes of the underlying graph. Local gradients are harnessed to further modulate message passing updates. Our framework flexibly allows one to use any basic GNN layer as a wrapper around which the multi-rate gradient gating mechanism is built. We rigorously prove that \model~alleviates the oversmoothing problem and allows the design of deep GNNs. Empirical results are presented to demonstrate that the proposed framework achieves state-of-the-art performance on a variety of graph learning tasks, including on large-scale heterophilic graphs.
\end{abstract}

\section{Introduction}
Learning tasks involving graph structured data arise in a wide variety of problems in science and engineering. 
Graph Neural Networks (GNNs) \citep{sperduti1994encoding,goller1996learning,sperduti1997supervised,frasconi1998general,gori2005new,scarselli2008graph,bruna2013spectral,chebnet,gcn,MoNet,mpnn} are a popular deep learning architecture for graph-structured and relational data. 
GNNs have been successfully applied in domains including computer vision and graphics \citep{MoNet}, recommender systems \citep{ying2018graph}, transportation \citep{derrowpinion2021traffic}, computational chemistry \citep{mpnn}, drug discovery \citep{gaudelet2021utilizing}, particle physics \citep{shlomi2020graph} and social networks.
See \citet{zhou,gdlbook} for extensive reviews.  

Despite the widespread success of GNNs and a plethora of different architectures, several fundamental problems still impede their efficiency on realistic learning tasks. 
These include the bottleneck  \citep{alon2020bottleneck},  oversquashing  \citep{topping2021understanding}, and oversmoothing  \citep{os1,os2} phenomena. 
Oversmoothing refers to the observation that all node features in a deep (multi-layer) GNN converge to the same constant value as the number of layers is increased. 
Thus, and in contrast to standard machine learning frameworks, oversmoothing inhibits the use of very deep GNNs for learning tasks. 
These phenomena are likely responsible for the unsatisfactory empirical performance of traditional GNN architectures in \emph{heterophilic} datasets, where the features or labels of a node tend to be different from those of its neighbors \citep{syn_cora}. 

Given this context, our main goal is to present a novel framework that  alleviates the oversmoothing problem and allows one to implement very deep multi-layer GNNs that can significantly improve performance in the setting of heterophilic graphs.
Our starting point is the observation that in standard Message-Passing GNN architectures (MPNNs), such as GCN \citep{gcn} or GAT \citep{gat}, each node gets updated at exactly the {\em same rate} within every hidden layer. 
Yet, realistic learning tasks might benefit from having different rates of propagation (flow) of information on the underlying graph. 
This insight leads to a novel \emph{multi-rate message passing} scheme capable of learning these underlying rates. 
Moreover, we also propose a novel procedure that harnesses graph gradients to ameliorate the oversmoothing problem. 
Combining these elements leads to a new architecture described in this paper, which we term \textbf{Gradient Gating} (\textbf{G$^2$}).  

\paragraph{Main Contributions.} 
We will demonstrate the following advantages of the proposed approach: 
\begin{itemize}[leftmargin=*]
\item 
\model~is a  flexible framework wherein any standard message-passing layer (such as GAT, GCN, GIN, or GraphSAGE) can be used as the  coupling function. Thus, it should be thought of as a framework into which one can plug existing GNN components. The use of multiple rates and gradient gating 
facilitates the implementation of deep GNNs and generally improves performance. 

\item 
\model~can be interpreted as a discretization of a dynamical system governed by nonlinear differential equations.  
By investigating the stability of zero-Dirichlet energy steady states of this system, we rigorously prove that our gradient gating mechanism  prevents oversmoothing. 
To complement this, we also prove a partial converse, that the lack of gradient gating can lead to oversmoothing.

\item 
We provide extensive empirical evidence demonstrating that \model~achieves state-of-the-art performance on a variety of graph learning tasks, including on large heterophilic graph~datasets. 
\end{itemize}

\section{Gradient Gating}
Let $\mathcal{G}=(\mathcal{V},\mathcal{E}\subseteq \mathcal{V}\times \mathcal{V})$ be an undirected graph with $|\mathcal{V}|=v$ nodes and $|\mathcal{E}|=e$ edges (unordered pairs of nodes $\{ i,j \}$   denoted $i \sim j$). 
The \emph{$1$-neighborhood} of a node $i$ is denoted  
$
\cN_i =
\{j \in \cV : i\sim j \}
$.
Furthermore, each node $i$ is endowed with an $m$-dimensional feature vector $\mathbf{X}_i$; the node features are arranged into a $v\times m$ matrix 
$\bX = (\bX_{ik})$ with $i=1,\hdots, v$ and $k=1,\hdots, m$.

A typical residual Message-Passing GNN (MPNN) updates the node features by performing several iterations of the form,
\begin{equation}
\label{eq:mp}
\bX^n = \bX^{n-1} + \sigma(\bF_\theta(\bX^{n-1},\cG)),
\end{equation}
where $\bF_\theta$ is a \emph{learnable} function  with parameters $\theta$, and $\sigma$ is an element-wise non-linear activation function. Here $n \geq 1$ denotes the $n$-th hidden layer with  $n=0$ being the input.

One can interpret \eqref{eq:mp} as a discrete dynamical system in which  $\mathbf{F}$ plays the role of a {\em coupling function} determining the interaction between different nodes of the graph. 
In particular, we consider local (1-neighborhood) coupling  of the form 
$\mathbf{Y}_i = (\mathbf{F}(\mathbf{X},\cG))_i = \mathbf{F}(\mathbf{X}_i, \ldblbrace  \mathbf{X}_{j\in \cN_i} \rdblbrace)$ operating on the multiset of 1-neighbors of each node. 
Examples of such functions used in the graph machine learning literature \citep{gdlbook} are {\em graph convolutions} $\mathbf{Y}_i = \sum_{j\in \cN_i} c_{ij} \mathbf{X}_j$ (GCN, \citep{gcn}) and {\em graph attention}  $ \mathbf{Y}_i= \sum_{j\in \cN_i} a(\mathbf{X}_i,\mathbf{X}_j) \mathbf{X}_j$ (GAT, \citep{gat}). 

\begin{figure}[ht]
\begin{center}
\begin{minipage}{.6\textwidth}
\begin{tikzpicture}
    \node[anchor=south west,inner sep=0] (image) at (0,0) {\includegraphics[width=1.\textwidth]{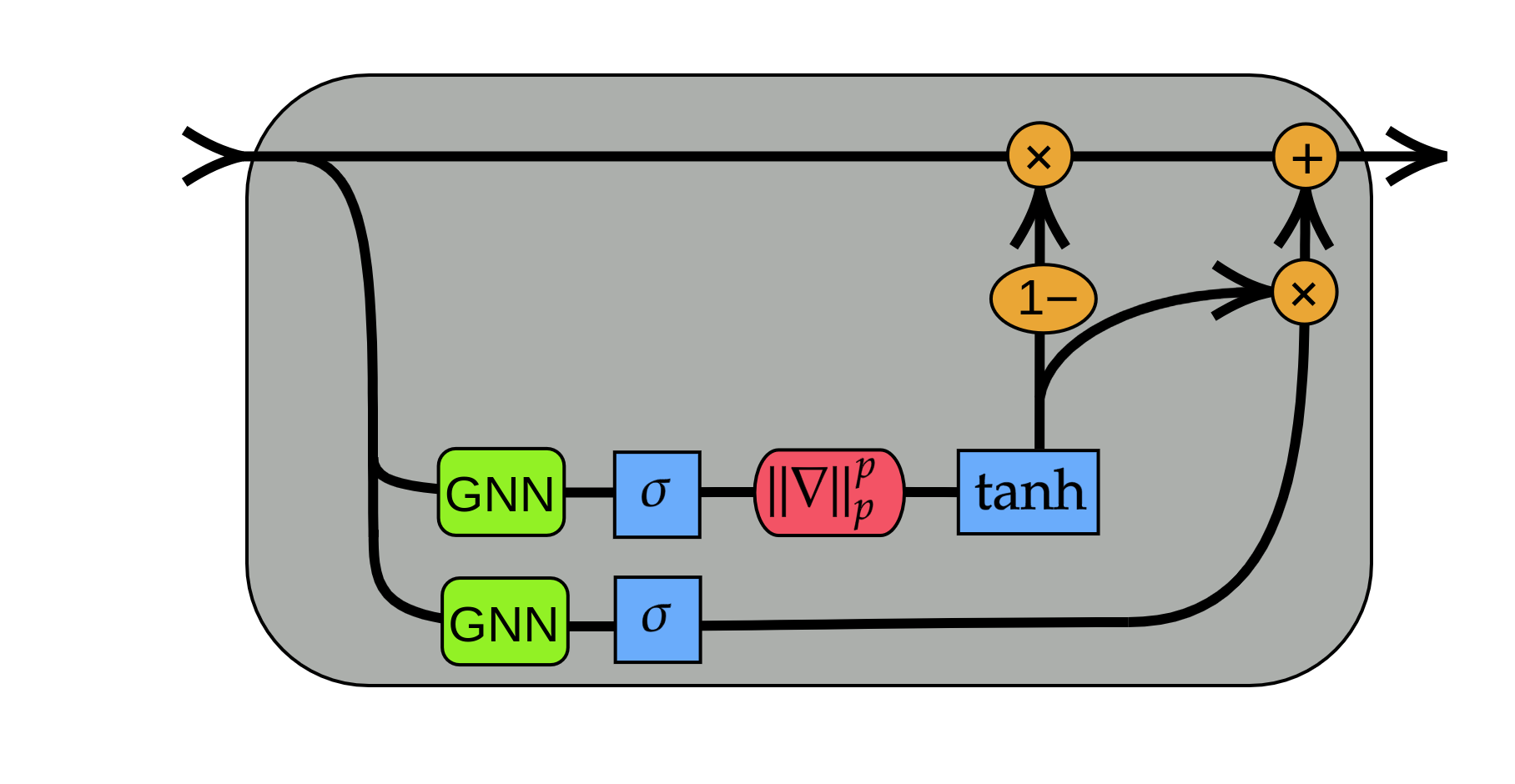}};
    \begin{scope}[x={(image.south east)},y={(image.north west)}]
        \draw (0.11, .85) node {$\bX^{n-1}$};
        \draw (0.98, .85) node {$\bX^{n}$};
        
    \end{scope}
\end{tikzpicture}
\caption{Schematic diagram of \model~\eqref{eq:ms_mp_non_oversmoothing} showing the layer-wise update of the latent node features $\bX$ (at layer $n$). The norm of the graph-gradient (i.e., sum in second equation in \eqref{eq:ms_mp_non_oversmoothing}) is denoted as $\|\nabla\|^p_p$.}
\label{fig:gradient_gating_scheme}
\end{minipage}
\end{center}
\end{figure}

We observe that in \eqref{eq:mp}, at each hidden layer, every node and every feature channel gets updated with exactly the same rate. 
However, it is reasonable to expect that in realistic graph learning tasks one can encounter multiple rates for the flow of information (node updates) on the graph. Based on this observation, we propose a {\bf multi-rate (MR)} generalization of \eqref{eq:mp}, allowing updates to each node of the graph and feature channel with different rates,  
\begin{equation}
\label{eq:mr_mp}
    \bX^n =  (1 - \boldsymbol{\tau}^n)\odot \bX^{n-1} + \boldsymbol{\tau}^n \odot \sigma(\bF_\theta(\bX^{n-1},\cG)) ,
\end{equation}
where $\boldsymbol{\tau}$ denotes a $v\times m$ matrix of rates with elements  $\boldsymbol{\tau}_{ik} \in [0,1]$. 
Rather than fixing $\boldsymbol{\tau}$ prior to training, we aim to  learn the different update rates based on the node data $\bX$ and the local structure of the underlying graph $\cG$, as follows 
\begin{equation}
\begin{aligned}
\label{eq:ms_mp_oversmoothing}
    \boldsymbol{\tau}^n(\bX^{n-1},\cG) &= \bar{\sigma}(\hat{\bF}_{\hat{\theta}}(\bX^{n-1},\cG)),
\end{aligned}
\end{equation}
where $\hat{\bF}_{\hat{\theta}}$ is another learnable 1-neighborhood coupling function, and $\bar{\sigma}$ is a sigmoidal logistic activation function to constrain the rates to lie within $[0,1]$.
Since the multi-rate message-passing scheme \eqref{eq:mr_mp} using \eqref{eq:ms_mp_oversmoothing} does not necessarily prevent oversmoothing (for any choice of the coupling function), we need to further constrain the rate matrix $\boldsymbol{\tau}^n$.
To this end, we note that the \emph{graph gradient} of scalar node features $\by$ on the underlying graph $\cG$ is defined as $(\nabla \by)_{ij} = \by_j - \by_i$ at the edge $i \sim j$ \citep{ops_graphs}. Next, we will use graph gradients to
obtain the proposed {\bf Gradient Gating} (\model)~framework given by
\begin{equation}
\begin{aligned}
\label{eq:ms_mp_non_oversmoothing}
    \hat{\boldsymbol{\tau}}^n &= {\sigma}(\hat{\bF}_\theta(\bX^{n-1},\cG)),\\
    \boldsymbol{\tau}^n_{ik} &= 
    \tanh\left(\sum_{j\in\cN_i}|\hat{\boldsymbol{\tau}}^n_{jk} - \hat{\boldsymbol{\tau}}^n_{ik}|^p\right), \\
    \bX^n &=  (1 - \boldsymbol{\tau}^n)\odot \bX^{n-1} + \boldsymbol{\tau}^n \odot \sigma(\bF_\theta(\bX^{n-1},\cG)),
\end{aligned}
\end{equation}
where $\hat{\boldsymbol{\tau}}^n_{jk} - \hat{\boldsymbol{\tau}}^n_{ik}=(\nabla \hat{\boldsymbol{\tau}}_{* k}^n)_{ij}$ denotes the graph-gradient and $\hat{\boldsymbol{\tau}}_{* k}^n$ is the $k$-th column of the rate matrix $\hat{\boldsymbol{\tau}}^n$ and $p\geq 0$. 
Since $\sum_{j\in\cN_i}|\hat{\boldsymbol{\tau}}^n_{jk} - \hat{\boldsymbol{\tau}}^n_{ik}|^p \geq 0$ for all $i\in \cV$, it follows that $\boldsymbol{\tau}^n \in [0,1]^{v \times m}$ for all $n$, retaining its interpretation as a matrix of rates. The sum over the neighborhood $\cN_i$ in \eqref{eq:ms_mp_non_oversmoothing} can be replaced by any permutation-invariant aggregation function (e.g., mean or max). 
Moreover, any standard message-passing procedure  can be used to define the coupling functions $\bF$ and  $\hat{\bF}$ (and, in particular, one can set $\hat{\bF} = \bF$). 
As an illustration, \fref{fig:gradient_gating_scheme} shows a schematic diagram of the layer-wise update of the proposed \model~architecture.

The intuitive idea behind gradient gating in \eqref{eq:ms_mp_non_oversmoothing} is the following:
If for any node $i\in \cV$ local oversmoothing occurs, i.e., $\lim_{n\to \infty} \sum_{j\in \cN_i}\|\bX^n_i - \bX^n_j \| = 0$, then \model~ensures that the corresponding rate $ \boldsymbol{\tau}^n_i$ goes to zero (at a faster rate), such that the underlying hidden node feature $\bX_i$ is no longer updated. 
This prevents oversmoothing by \emph{early-stopping} of the message passing procedure.   

\section{Properties of \model-GNN}

\paragraph{\model~is a flexible framework.} 
An important aspect of \model~\eqref{eq:ms_mp_non_oversmoothing} is that it can be considered as a ``wrapper'' around any specific MPNN architecture. 
In particular, the hidden layer update for \emph{any} form of message passing (e.g., GCN \citep{gcn}, GAT \citep{gat}, GIN \citep{gin} or GraphSAGE \citep{graphsage}) can be used as the coupling functions $\bF,\hat{\bF}$ in  \eqref{eq:ms_mp_non_oversmoothing}. 
By setting $\btau \equiv \mathbf{I}$,  \eqref{eq:ms_mp_non_oversmoothing} reduces to\vspace{-0.5mm} 
\begin{equation}
\label{eq:sgnn}
\bX^n =  \sigma\left(\bF_\theta(\bX^{n-1},\cG)\right),
\end{equation}
a standard (non-residual) MPNN. 
As we will show in the following, the use of a non-trivial gradient-gated \emph{learnable} rate matrix $\btau$ allows implementing very deep architectures that avoid oversmoothing.

\paragraph{Maximum Principle for node features.}
Node features produced by \model~satisfy the following Maximum Principle.
\begin{proposition}
Let $\bX^n$ be the node feature matrix generated by iteration formula \eqref{eq:ms_mp_non_oversmoothing}. Then, the features are bounded as follows:\vspace{-0.5mm}
\begin{equation}
    \min\left(-1,\underline{\sigma}\right) \leq \bX^n_{ik} \leq \max\left(1,\overline{\sigma}\right),\quad \forall 1\leq n,
\end{equation}
where the scalar activation function is bounded by 
$
\underline{\sigma} \leq \sigma (z) \leq \overline{\sigma} 
$
for all $z \in \R$.
\end{proposition}
The proof follows readily from writing \eqref{eq:ms_mp_non_oversmoothing} component-wise and using the fact that $0 \leq \boldsymbol{\tau}^n_{ik} \leq 1$, for all $1 \leq i \leq v$, $1 \leq k \leq m$ and $1\leq n$.

\paragraph{Continuous limit of \model.}
It has recently been shown (see \citet{Ave1,Poli2019,Zhu1,Xhonneux2020,grand,pde-gcn,blend,topping2021understanding,graphcon} and references therein) that interesting properties of GNNs (with residual connections) can be understood by taking the continuous (infinite-depth) limit and analyzing the resulting 
differential equations.

In this context, we can derive a continuous version of \eqref{eq:ms_mp_non_oversmoothing} by introducing a small-scale $0 < \Delta t < 1$ and rescaling the rate matrix $ \boldsymbol{\tau}^n$ to $\Delta t  \boldsymbol{\tau}^n$ leading to
\begin{equation}
    \label{eq:ggrs}
    \bX^n =  (1 - \Delta t \boldsymbol{\tau}^n)\odot \bX^{n-1} + \Delta t \boldsymbol{\tau}^n \odot \sigma\left(\bF_\theta(\bX^{n-1},\cG)\right) .
\end{equation}
Rearranging the terms in \eqref{eq:ggrs}, we obtain
\begin{equation}
    \label{eq:ggrs1}
     \frac{\bX^n - \bX^{n-1}}{\Delta t} = \boldsymbol{\tau}^n \odot \left(
    \sigma\left(\bF_\theta(\bX^{n-1},\cG)\right) - \bX^{n-1}\right) .
\end{equation}
Interpreting $\bX^n \approx \bX(n \Delta t) = \bX(t^n)$, i.e., marching in time, corresponds to increasing the number of hidden layers.
Letting $\Delta t \rightarrow 0$, one obtains the following system of graph-coupled ordinary differential equations (ODEs):
\begin{equation}
    \label{eq:ggode}
    \begin{aligned}
    \frac{d\bX(t)}{dt} &= \btau(t) \odot \left(
    \sigma\left(\bF_\theta(\bX(t),\cG)\right) - \bX(t)\right), \quad \forall t\geq 0, \\
    \boldsymbol{\tau}_{ik}(t) &= \tanh\left(\sum_{j\in\cN_i}|\hat{\boldsymbol{\tau}}_{ik}(t) - \hat{\boldsymbol{\tau}}_{jk}(t)|^p\right), 
    \\ 
    \hat{\boldsymbol{\tau}}(t) &= \hat{\sigma}(\hat{\bF}_{\hat{\theta}}(\bX^{n-1},\cG) ).
    \end{aligned}
\end{equation}
We observe that the iteration formula  \eqref{eq:ms_mp_non_oversmoothing} acts as a \emph{forward Euler} discretization of the ODE system \eqref{eq:ggode}. 
 Hence, one can follow \citet{grand} and design more general (e.g., higher-order, adaptive, or implicit) discretizations of the ODE system \eqref{eq:ggode}. All these can be considered as design extensions of \eqref{eq:ms_mp_non_oversmoothing}. 

\paragraph{Oversmoothing.} 
Using the interpretation of \eqref{eq:ms_mp_non_oversmoothing} as a discretization of the ODE system \eqref{eq:ggode}, we can adapt the mathematical framework recently proposed in \citet{graphcon} to study the oversmoothing problem. 
In order to formally define oversmoothing, we introduce the \emph{Dirichlet energy} 
defined on the node features $\bX$ of an undirected graph $\cG$ as 
\begin{equation}
\label{eq:graph_H1}
    \cE(\bX) = \frac{1}{v}\sum_{i \in \cV} \sum_{j \in \cN_i} \|\mathbf{X}_i - \mathbf{X}_j\|^2.\vspace{-1mm}
\end{equation}
Following \citet{graphcon}, we say that the scheme \eqref{eq:ggode} {\em oversmoothes} if the 
Dirichlet energy decays {\em exponentially fast}, %\vspace{-1mm}
\begin{equation}
    \label{eq:vgg1}
\cE(\bX(t)) \leq C_1 e^{-C_2t}, \quad \forall t > 0,%\vspace{-1mm}
\end{equation}
for some $C_{1,2} > 0$. In particular, the discrete version of \eqref{eq:vgg1} implies that oversmoothing happens when the Dirichlet energy, decays exponentially fast as the number of hidden layers increases (\citep{graphcon} Definition 3.2).

Next, one can prove the following proposition further characterizing oversmoothing with the standard terminology of dynamical systems \citep{Wig1}.
\begin{proposition}
\label{prop:2}
The oversmoothing problem occurs for the ODEs \eqref{eq:ggode} iff the hidden states $\bX^\ast_i = \bc$, for all $i \in \cV$ are \emph{exponentially stable steady states (fixed points)} of the ODE \eqref{eq:ggode}, for some $\bc \in \R^{m}$. 
\end{proposition}
In other words, for the oversmoothing problem to occur for this system, all the trajectories of the ODE \eqref{eq:ggode} that start within the corresponding basin of attraction have to converge exponentially fast in time  (according to \eqref{eq:vgg1}) to the corresponding steady state $\bc$.
Note that the basins of attraction will be different for different values of $\bc$. 
The proof of this Proposition is a straightforward adaptation of the proof of Proposition 3.3 of \citet{graphcon}. 

Given this precise formulation of oversmoothing, we will investigate whether and how gradient gating in  \eqref{eq:ggode} can prevent oversmoothing. 
For simplicity, we set $m=1$ to consider only scalar node features (extension to vector node features is straightforward). Moreover, we assume coupling functions of the form $\bF(\bX) = \mathbf{A}(\mathbf{X})\mathbf{X}$, expressed element-wise as (see also \citet{grand,graphcon}),
\begin{equation}
    (\bF(\bX))_i = \sum_{j \in \cN_i} \bA(\mathbf{X}_i,\mathbf{X}_j) \mathbf{X}_j. 
\end{equation}
Here, $\mathbf{A}(\mathbf{X})$ is a matrix-valued function whose form covers many commonly used coupling functions stemming from the graph attention (GAT, where $\bA_{ij} = \bA(\mathbf{X}_i,\mathbf{X}_j)$ is learnable) or convolution operators (GCN, where $\bA_{ij}$ is fixed). 
Furthermore, the matrices are \emph{right stochastic}, i.e., the entries satisfy
\begin{equation}
\label{eq:aij}
\begin{aligned}
0 \leq \bA_{ij} &\leq 1,~\sum\limits_{j\in\cN_i} \bA_{ij}&= 1. 
\end{aligned}
\end{equation}
Finally, as the multi-rate feature of \eqref{eq:ggode} has no direct bearing on the oversmoothing problem, we focus on the contribution of the gradient feedback term. To this end, we deactivate the multi-rate aspects and assume that $\hat{\btau}_i = \mathbf{X}_i$ for all $i \in \cV$, leading to the following form of \eqref{eq:ggode}:
\begin{equation}
    \label{eq:ggode1}
    \begin{aligned}
      \frac{d\mathbf{X}_i(t)}{dt} &= \btau_i(t) \left(
    \sigma\left( \sum_{j \in \cN_i} \bA_{ij} \mathbf{X}_j(t) \right) - \mathbf{X}_i(t) \right), \quad \forall t\geq 0,  
    \\ 
    \btau_{i}(t) &= \tanh\left(\sum_{j\in\cN_i}\|\mathbf{X}_{j}(t) - \mathbf{X}_{i}(t)\|^p\right). 
    \end{aligned}
\end{equation}

\paragraph{Lack of \model~can lead to oversmoothing.} 
We first consider the case where the Gradient Gating is switched off by setting $p=0$ in \eqref{eq:ggode1}. 
This yields a standard GNN in which node features are evolved through message passing between neighboring nodes, without any explicit information about graph gradients. We further assume that the activation function is ReLU i.e., $\sigma(x) = \max(x,0)$. Given this setting, we have the following proposition on oversmoothing: 
\begin{proposition}
\label{prop:3}
Assume the underlying graph $\cG$ is  connected. For any $c \geq 0$, let  $\bX^{\ast}_i \equiv c$, for all $i \in \cV$ be a (zero-Dirichlet energy) steady state of the ODEs \eqref{eq:ggode1}. Moreover, 
assume no Gradient Gating ($p=0$ in \eqref{eq:ggode1}) and  
\begin{equation}
    \label{eq:assm2}
    \bA_{ij}(c,c) = \bA_{ji}(c,c), ~ {\rm and}~ \bA_{ij}(c,c) \geq \underline{a}, \quad 1 \leq i,j \leq v,
\end{equation}
with $0 < \underline{a} \leq 1$
and that there exists at least one node denoted w.l.o.g. with index $1$ such that $\mathbf{X}_1(t) \equiv c$, for all $t\geq 0$. Then, the steady state $\bX^{\ast}_i = \bc$, for all $i \in \cV$, of \eqref{eq:ggode1} is \emph{exponentially stable}. 
\end{proposition}
Proposition \ref{prop:2} implies that without gradient gating (\model), \eqref{eq:ggode} can lead to oversmoothing. The proof, presented in {\bf SM} \ref{app:prop3pf} relies on analyzing the time-evolution of small perturbations around the steady state $\bc$ and showing that these perturbations decay exponentially fast in time (see  \eqref{eq:pf3}).

\paragraph{\model~prevents oversmoothing.} 
We next investigate the effect of Gradient Gating  in the same setting of Proposition \ref{prop:3}. 
The following Proposition shows that gradient gating prevents oversmoothing: 
\begin{proposition}
\label{prop:4}
Assume the underlying graph $\cG$ is connected. For any $c \geq 0$ and for all $i \in \cV$, let  $\bX_i^{\ast} \equiv c$ be a (zero-Dirichlet energy) steady state of the ODEs \eqref{eq:ggode1}. Moreover, assume  Gradient Gating ($p>0$)  and that the matrix $\bA$ in \eqref{eq:ggode1} satisfies \eqref{eq:assm2} and that there exists at least one node denoted w.l.o.g. with index $1$ such that $\bX_1(t) \equiv c$, for all $t\geq 0$. Then, the steady state $\bX^{\ast}_i = \bc$, for all $i \in \cV$ is \emph{not exponentially stable}. 
\end{proposition}

The proof, presented in {\bf SM} \ref{app:prop4pf} clearly elucidates the role of gradient gating by showing that the energy associated with the quasi-linearized evolution equations ({\bf SM} Eqn. \eqref{eq:ggode4}) is balanced by two terms ({\bf SM} Eqn. \eqref{eq:pf5}), both resulting from the introduction of gradient gating by setting $p > 0$ in \eqref{eq:ggode1}. 
One of them is of indefinite sign and can even cause \emph{growth} of perturbations around a steady state $\bc$. 
The other decays initial perturbations. 
However, the rate of this decay is at most \emph{polynomial} ({\bf SM} Eqn. \eqref{eq:pf10}). 
For instance, the decay is merely linear for $p=2$ and slower for higher values of $p$. 
Thus, the steady state $\bc$ cannot be exponentially stable and oversmoothing is prevented. 
This justifies the intuition behind gradient gating, namely, if oversmoothing occurs around a node $i$, i.e., $\textstyle \lim_{n\to \infty} \sum_{j\in \cN_i}\|\bX^n_i - \bX^n_j \| = 0$, then the corresponding rate $ \boldsymbol{\tau}^n_i$ goes to zero (at a faster rate), such that the underlying hidden node feature $\bX_i$ stops getting updated.

\section{Experimental Results}

In this section, we present an experimental study of  \model~on both synthetic and real datasets.  
We use \model~with three different coupling functions: GCN \citep{gcn}, GAT \citep{gat} and GraphSAGE \citep{graphsage}.

\begin{figure}[ht!]
\centering
%\vspace{-7mm}
\begin{minipage}[t]{.475\textwidth}
\includegraphics[width=1.\textwidth]{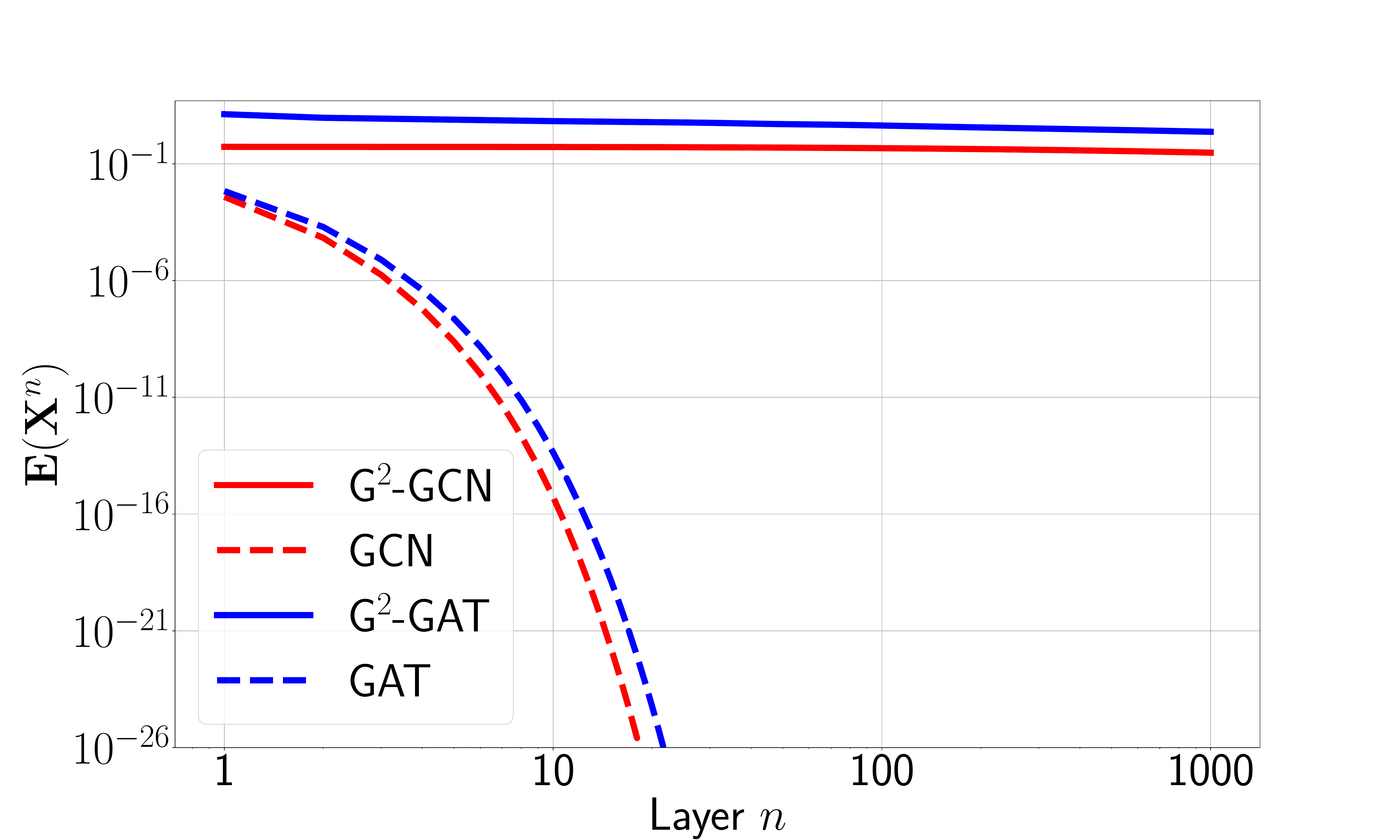}
\caption{Dirichlet energy $\cE(\bX^n)$ of layer-wise node features $\bX^n$ propagated through a GAT, GCN and their gradient gated versions (\model-GAT, \model-GCN).}
\label{fig:dirichlet_plot}
\end{minipage}%
\hfill
\begin{minipage}[t]{.475\textwidth}
\includegraphics[width=1.\textwidth]{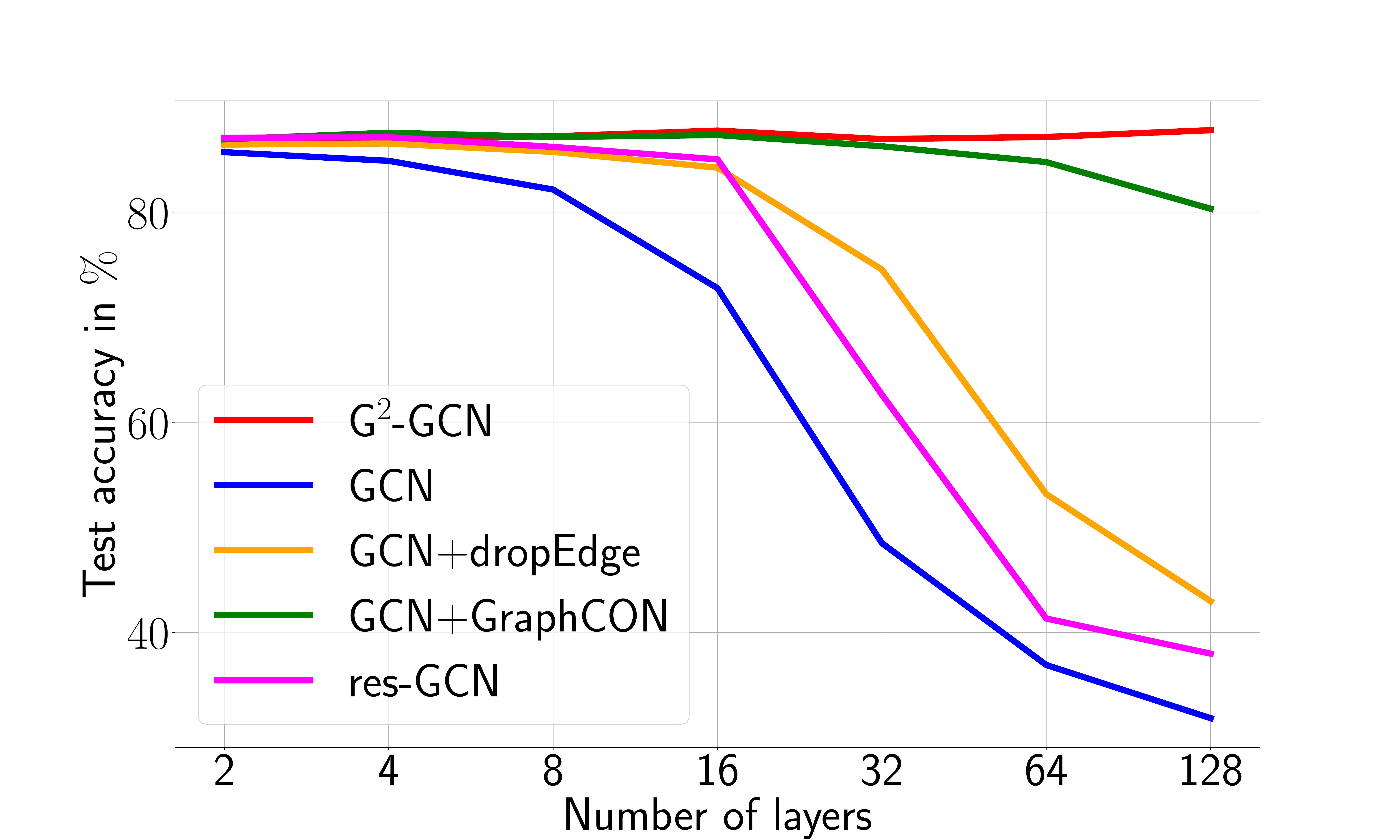}
\caption{Test accuracies of GCN with gradient gating (\model-GCN) as well as plain GCN and GCN combined with other methods on the Cora dataset for increasing number of layers.}
\label{fig:depth_vs_acc}
\end{minipage}%
\vspace{-3mm}
\end{figure}

\paragraph{Effect of \model~on Dirichlet energy.}
Given that oversmoothing relates to the decay of Dirichlet energy \eqref{eq:vgg1}, we follow the experimental setup proposed by \citet{graphcon} to probe the dynamics of the Dirichlet energy of Gradient-Gated GNNs, defined on a 
2-dimensional $10\times10$ regular grid
with 4-neighbor connectivity.
The node features $\bX$ are randomly sampled from $\mathcal{U}([0,1])$ and then propagated through a $1000$-layer GNN with random weights. We compare GAT, GCN and their gradient-gated versions (\model-GAT and \model-GCN) in this experiment. 
\fref{fig:dirichlet_plot} depicts on log-log scale the 
Dirichlet energy 
of each layer's output with respect to the layer number. We clearly observe that GAT and GCN \emph{oversmooth} as the underlying Dirichlet energy converges exponentially fast to zero, resulting in the node features becoming indistinguishable. 
In practice, the Dirichlet energy for these architectures is $\approx 0$ after just ten hidden layers. 
On the other hand, and as suggested by the theoretical results of the previous section, adding \model~decisively prevents this behavior and the Dirichlet energy remains (near) constant, even for very deep architectures (up to 1000 layers). 

\paragraph{\model~for very deep GNNs.}
Oversmoothing inhibits the use of large number of GNN layers. As \model~is designed to alleviate oversmoothing, it should allow  very deep architectures. To test this assumption, we reproduce the experiment considered in \citet{grand}: a node-level classification task on the Cora dataset using increasingly deeper GCN architectures.  
In addition to \model, we also compare with two recently proposed mechanisms to alleviate oversmoothing, DropEdge \citep{dropedge} and GraphCON \citep{graphcon}. 
The results are presented in \fref{fig:depth_vs_acc}, where we plot the test accuracy for all the models with the number of layers ranging from $2$ to $128$. 
While a plain GCN seems to suffer the most from oversmoothing (with the performance rapidly deteriorating after 8 layers), GCN+DropEdge as well as GCN+GraphCON are able to mitigate this behavior to some extent, although the performance eventually starts dropping (after 16 and 64  layers, respectively). 
In contrast, \model-GCN exhibits a small but noticeable \emph{increase} in performance for increasing number of layers, reaching its peak performance for $128$ layers. 
This experiment suggests that \model~can indeed be used in conjunction with deep GNNs, potentially allowing performance gains due to depth. 

\begin{figure}[ht!]
\vspace{-2mm}
\centering
\begin{minipage}[t]{.475\textwidth}
\includegraphics[width=1.\textwidth]{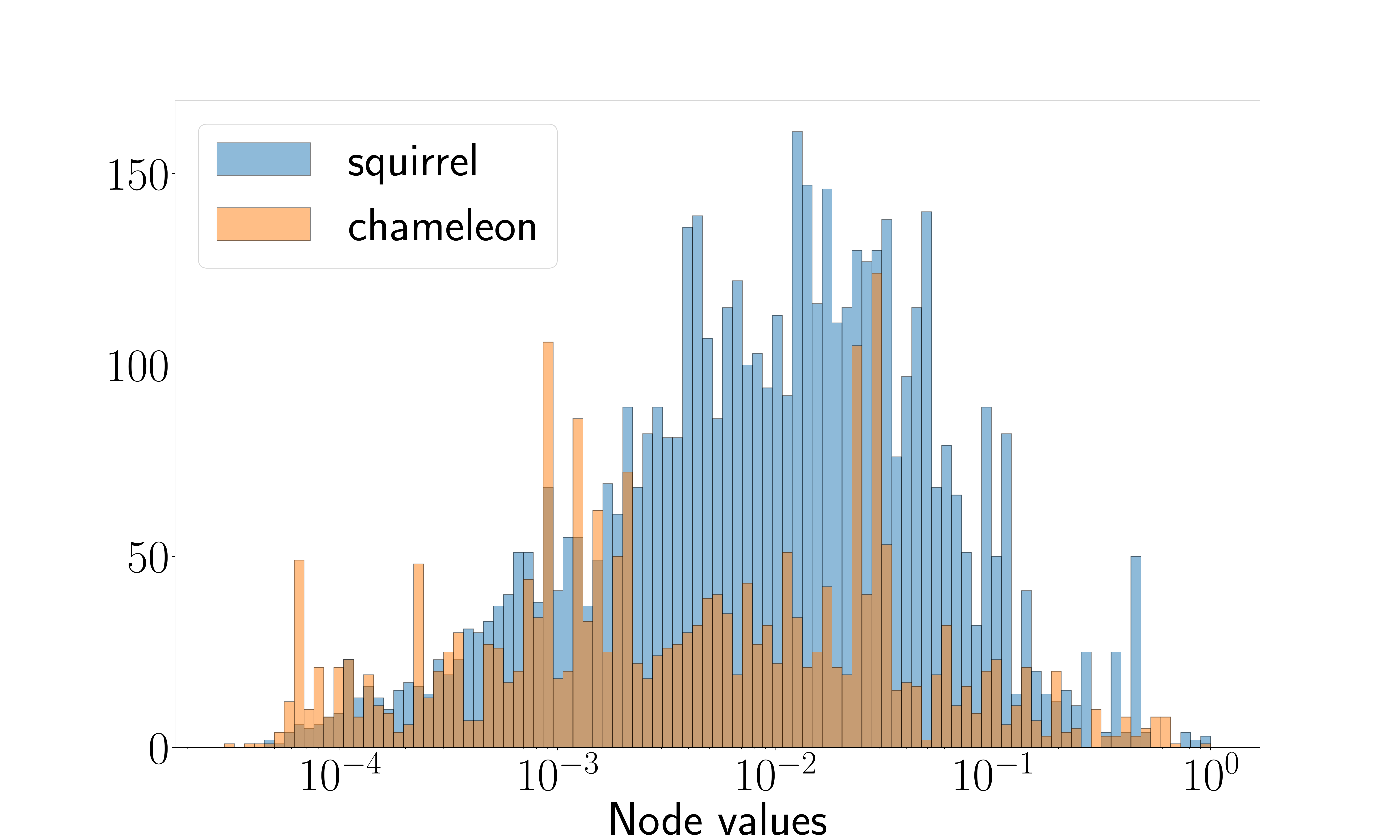}
\caption{Histogram of the target node values of the Chameleon and Squirrel node-level regression tasks.}
\label{fig:wiki_reg_hist}
\end{minipage}%
\hfill
\begin{minipage}[t]{.475\textwidth}
\includegraphics[width=1.\textwidth]{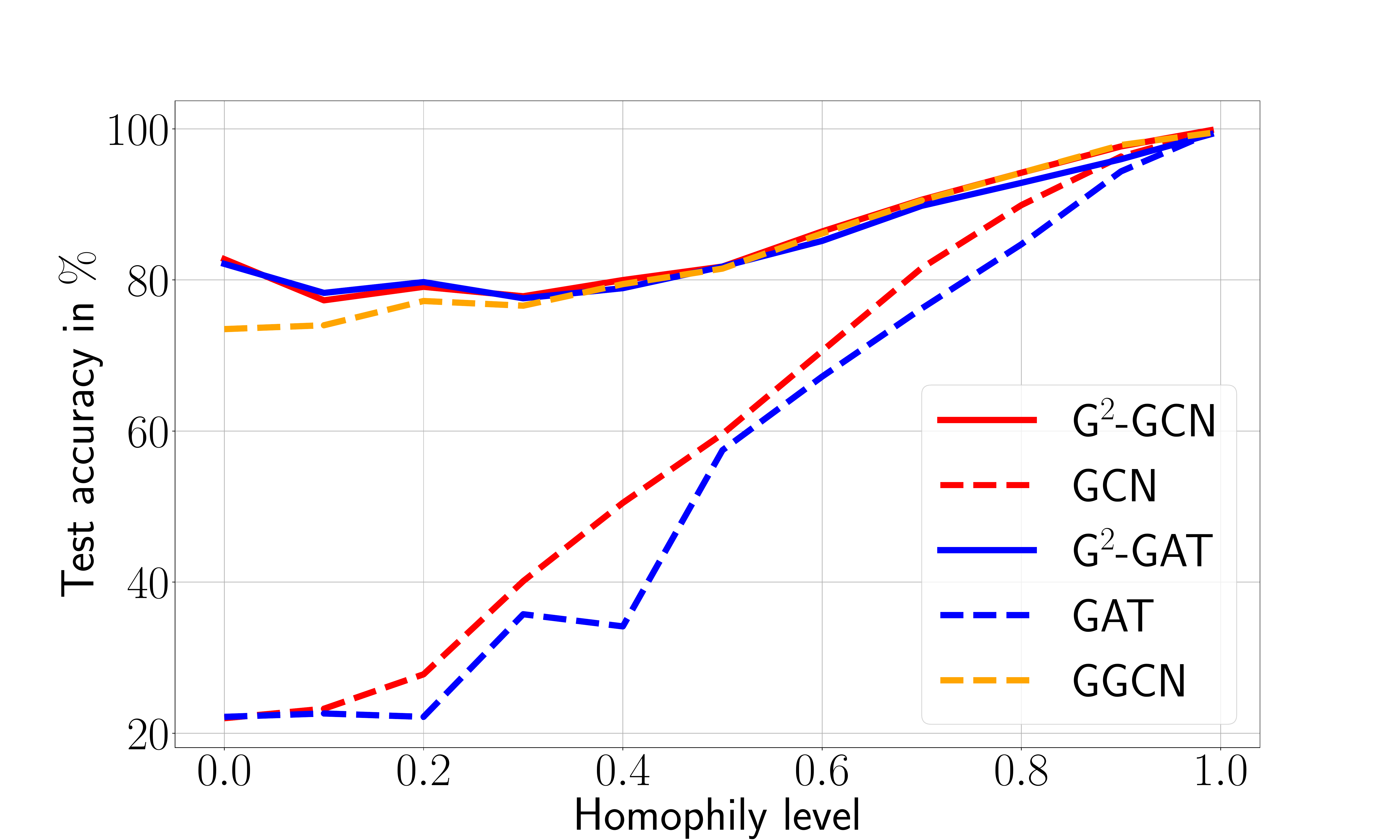}
\caption{Test accuracy of GCN and GAT with / without gradient gating (\model) on synthetic Cora with a varying level of  true label homophily.}
\label{fig:syn_cora}
\end{minipage}%
\vspace{-3mm}
\end{figure}

\paragraph{\model~for multi-scale node-level regression.}
\begin{wraptable}{r}{0.4\textwidth}
    \centering
    \caption{Normalized test MSE on multi-scale node-level regression tasks.}
    \label{tab:node_reg}
    \resizebox{0.4\textwidth}{!}{
    \begin{tabular}{l cc}
    \toprule 
         &
         \textbf{Chameleon} &
         \textbf{Squirrel} \\
         
         \#Nodes &
         2,277 &
         5,201 \\
        
        \#Edges &
        31,421& 
        198,493
         \\ \midrule
        
        GCNII &
        $\bf \color{violet} 0.170 \pm 0.034$ &
        $\bf \color{violet} 0.093 \pm 0.031$ \\
        
        PairNorm &
        $0.207 \pm 0.038$ &
        $0.140 \pm 0.040$
        \\
        
        GCN &
        $0.207 \pm 0.039$&
        $0.143 \pm 0.039$
        \\
        
        GAT & 
        $0.207 \pm 0.038$&
        $0.143 \pm 0.039$
        \\
        
        \midrule
        \model-GCN & 
        $\bf \color{blue} 0.137 \pm 0.033$&
        $\bf \color{blue} 0.070 \pm 0.028$ \\
        
        \model-GAT & 
        $\bf \color{red} 0.136 \pm 0.029$&
        $\bf \color{red} 0.069 \pm 0.029$ \\
        
        \bottomrule
        \vspace{-3mm}
    \end{tabular}
    }
\end{wraptable} 
We test the multi-rate  nature of  \model~on node-level regression tasks, where the target node values exhibit multiple scales. 
Due to a lack of widely available node-level regression tasks, we propose regression experiments based on the Wikipedia article networks  Chameleon and Squirrel, \citep{wiki_datasets}. While Chameleon and Squirrel are already widely used as heterophilic node-level classification tasks, the original datasets consist of continuous node targets (average monthly web-page traffic). We normalize the provided webpage traffic values for every node  between 0 and 1 and note that the resulting node values exhibit values on a wide range of different scales ranging between $10^{-5}$ and $1$ (see \fref{fig:wiki_reg_hist}). 
\Tref{tab:node_reg} shows the test normalized mean-square error (mean and standard deviation based on the ten pre-defined splits in \citet{geom_gcn}) for two standard GNN architectures (GCN and GAT) with and without \model. We observe from \Tref{tab:node_reg} that adding \model~to the baselines significantly reduces the error, demonstrating the advantage of using multiple update rates.

\paragraph{\model~for varying homophily (Synthetic Cora).}
We test \model~on a node-level classification task with varying levels of homophily on the synthetic Cora dataset \citet{syn_cora}. 
Standard GNN models are known to perform poorly in heterophilic settings. 
This can be seen in \fref{fig:syn_cora}, where we present the classification accuracy of GCN and GAT on the synthetic-Cora dataset with a  level of homophily varying between $0$ and $0.99$. While these models succeed in the homophilic case (reaching nearly perfect accuracy), their performance drops to $\approx 20\%$ when the level of homophily approaches $0$. 
Adding \model~to GCN or GAT mitigates this phenomenon: 
the resulting models reach a test accuracy of over $80\%$, even in the most heterophilic setting, thus leading to a four-fold increase in the accuracy of the underlying GCN or GAT models. Furthermore, we notice an increase in performance even in the homophilic setting. 
Moreover, we compare with a state-of-the-art model GGCN \citep{ggcn}, which has been recently proposed to explicitly deal with heterophilic graphs. From \fref{fig:syn_cora} we observe that \model~performs on par and slightly better than GGCN in strongly heterophilic settings.

\begin{table*}[h!]
    \centering
    \caption{Results on heterophilic graphs. The three best performing methods are highlighted in {\bf \textcolor{red}{red}} (First), {\bf \textcolor{blue}{blue}} (Second), and {\bf \textcolor{violet}{violet}} (Third).%\vspace{-1mm}
    }
    \label{tab:hetero_tasks}
    \resizebox{\textwidth}{!}{%
    \begin{tabular}{l cccccc}
    \toprule 
         &
         \textbf{Texas} &  
         \textbf{Wisconsin} & 
         \textbf{Film} &
         \textbf{Squirrel} &
         \textbf{Chameleon} &
         \textbf{Cornell} \\
        
         Hom level &
         \textbf{0.11} &
         \textbf{0.21} & 
         \textbf{0.22} & 
         \textbf{0.22} & 
         \textbf{0.23} &
         \textbf{0.30} \\ 
         
         \#Nodes &
         183 &
         251 & 
         7,600 &
         5,201 & 
         2,277 &
         183 \\
        
        \#Edges &
        295 &
        466 & 
        26,752 & 
        198,493 & 
        31,421 &
        280 \\
        
        \#Classes &
        5 &
        5 & 
        5 &
        5 &
        5 & 
        5 \\ \midrule
        
        GGCN &
        $\bf \color{blue} 84.86 \pm 4.55$ &
        $86.86 \pm 3.29$ &
        $\bf \color{red} 37.54 \pm 1.56$ &
        $55.17 \pm 1.58$ & 
        $\bf \color{blue} 71.14 \pm 1.84$ &
        $85.68 \pm 6.63$ \\

        GPRGNN &
        $78.38 \pm 4.36$ &
        $82.94 \pm 4.21$ &
        $34.63 \pm 1.22$ & 
        $31.61 \pm 1.24$ &
        $46.58 \pm 1.71$ &
        $80.27 \pm 8.11$ \\ 

        H2GCN &
        $\bf \color{blue} 84.86 \pm 7.23$ &
        $\bf \color{blue} 87.65 \pm 4.98$ &
        $35.70 \pm 1.00$ &
        $36.48 \pm 1.86$ & 
        $60.11 \pm 2.15$ &
        $82.70 \pm 5.28$ \\
        
        FAGCN &
        $82.43 \pm 6.89$ & 
        $82.94 \pm 7.95$ &
        $34.87 \pm 1.25$ & 
        $42.59 \pm 0.79$ &
        $55.22 \pm 3.19$ &
        $79.19 \pm 9.79$ \\
        
        F$^2$GAT &
        $82.70 \pm 5.95$ &
        $\bf \color{violet} 87.06 \pm 4.13$ &
        $36.65 \pm 1.13$ &
        $47.32 \pm 2.43$ &
        $67.81 \pm 2.05$ &
        $83.51 \pm 6.70$
        \\

        MixHop &
        $77.84 \pm 7.73$ &
        $75.88 \pm 4.90$ &
        $32.22 \pm 2.34$ &
        $43.80 \pm 1.48$ &
        $60.50 \pm 2.53$ &
        $73.51 \pm 6.34$ \\

        GCNII &
        $77.57 \pm 3.83$ &
        $80.39 \pm 3.40$  &
        $\bf \color{blue} 37.44 \pm 1.30$ &
        $38.47 \pm 1.58$ &
        $63.86 \pm 3.04$ &
        $77.86 \pm 3.79$ \\
        
        Geom-GCN &
        $66.76 \pm 2.72$ &
        $64.51 \pm 3.66$ &
        $31.59 \pm 1.15$ & 
        $38.15 \pm 0.92$ & 
        $60.00 \pm 2.81$ &
        $60.54 \pm 3.67$ \\ 
        
        PairNorm &
        $60.27 \pm 4.34$ &
        $48.43 \pm 6.14$ &
        $27.40 \pm 1.24$ & 
        $50.44 \pm 2.04$ & 
        $62.74 \pm 2.82$ &
        $58.92 \pm 3.15$ \\ 
        
        LINKX &
        $74.60 \pm 8.37$ &
        $75.49 \pm 5.72$ &
        $36.10 \pm 1.55$ &
        $\bf \color{blue} 61.81 \pm 1.80$ &
        $68.42 \pm 1.38$ &
        $77.84 \pm 5.81$ \\
        
        GloGNN &
        $84.32 \pm 4.15$ &
        $\bf \color{violet} 87.06 \pm 3.53$ &
        $\bf \color{violet} 37.35 \pm 1.30$ &
        $\bf \color{violet} 57.54 \pm 1.39$ &
        $\bf \color{violet} 69.78 \pm 2.42$ &
        $83.51 \pm 4.26$ \\
        
        GraphSAGE &
        $82.43 \pm 6.14$ &
        $81.18 \pm 5.56$ &
        $34.23 \pm 0.99$ & 
        $41.61 \pm 0.74$ & 
        $58.73 \pm 1.68$ &
        $75.95 \pm 5.01$ \\
        
        ResGatedGCN &
        $80.00 \pm 5.57$ &
        $81.57 \pm 5.35$ &
        $36.02 \pm 1.19$ &
        $37.60 \pm 1.80$ &
        $49.82 \pm 2.71$ &
        $73.51 \pm 4.95$ \\
        
        GCN &
        $55.14 \pm 5.16$ &
        $51.76 \pm 3.06$ &
        $27.32 \pm 1.10$ & 
        $31.52 \pm 0.71$ &
        $38.44 \pm 1.92$ &
        $60.54 \pm 5.30$ \\ 
        
        GAT &
        $52.16 \pm 6.63$ &
        $49.41 \pm 4.09$ &
        $27.44 \pm 0.89$ & 
        $36.77 \pm 1.68$ &
        $48.36 \pm 1.58$ &
        $61.89 \pm 5.05$ \\ 
        
        MLP &
        $80.81 \pm 4.75$ &
        $85.29 \pm 3.31$ &
        $36.53 \pm 0.70$ & 
        $28.77 \pm 1.56$ & 
        $46.21 \pm 2.99$ &
        $81.89 \pm 6.40$ \\ \midrule
        
        \model-GAT & 
        $\bf \color{violet} 84.59 \pm 5.55$ &
        $\bf \color{blue} 87.65 \pm 4.64$ &
        $37.30 \pm 0.87$ & 
        $46.48 \pm 1.41$ & 
        $64.12 \pm 1.96$ &
        $\bf \color{red} 87.30 \pm 4.84$ \\
        
        \model-GCN & 
        $\bf \color{blue} 84.86 \pm 3.24$ &
        $\bf \color{violet} 87.06 \pm 3.19$ &
        $37.09 \pm 1.16$ & 
        $39.62 \pm 2.91$ & 
        $55.83 \pm 2.88$ &
        $\bf \color{blue} 86.49 \pm 5.27$ \\
        
        \model-GraphSAGE & 
        $\bf \color{red} 87.57 \pm 3.86$ &
        $\bf \color{red} 87.84 \pm 3.49$ &
        $37.14 \pm 1.01$ & 
        $\bf \color{red} 64.26 \pm 2.38$ & 
        $\bf \color{red} 71.40 \pm 2.38$ & 
        $\bf \color{violet} 86.22 \pm 4.90$ \\
        
        \bottomrule
    \end{tabular}
    }\vspace{-2mm}
\end{table*}

\paragraph{Heterophilic datasets.}
In \Tref{tab:hetero_tasks}, we test the proposed framework on several real-world heterophilic graphs (with a homophily level of $\leq 0.30$) \citep{geom_gcn,wiki_datasets} and benchmark it against baseline models GraphSAGE \citep{graphsage}, GCN \citep{gcn}, GAT \citep{gat} and MLP \citep{dlbook}, as well as recent state-of-the-art models on heterophilic graph datasets, i.e., GGCN \citep{ggcn}, GPRGNN \citep{gpr_gnn}, H2GCN \citep{syn_cora}, FAGCN \citep{fagcn}, F$^2$GAT \citep{f2gnn}, MixHop \citep{mixhop}, GCNII \citep{gcnii}, Geom-GCN \citep{geom_gcn}, PairNorm \citep{pairnorm}. 
We can observe that \model~added to GCN, GAT or GraphSAGE outperforms all other methods (in particular recent methods such as GGCN, GPRGNN, H2GCN that were explicitly designed to solve heterophilic tasks).
Moreover, adding \model~to the underlying base GNN model improves the results on average by $45.75\%$ for GAT, $45.4\%$ for GCN and $18.6\%$ for GraphSAGE.

\paragraph{Large-scale graphs.}
\begin{wraptable}{r}{0.59\textwidth}
\vspace{-5mm}
    \centering
    \caption{Results on large-scale datasets.\vspace{-2mm} 
    }
    \label{tab:hetero_large}
    \resizebox{0.59\textwidth}{!}{
    \begin{tabular}{l ccc}
    \toprule 
         &
         \textbf{snap-patents} &
         \textbf{arXiv-year} &
         \textbf{genius} \\
        
         Hom level &
         \textbf{0.07} &
         \textbf{0.22} &
         \textbf{0.61}\\ 
         
         \#Nodes &
         2,923,922 &
         169,343 &
         421,961 \\
        
        \#Edges &
        13,975,788& 
        1,166,243 &
        984,979 \\
        
        \#Classes &
        5 &
        5 &
        2 \\ \midrule
        
        MLP &
        $31.34 \pm 0.05$ &
        $36.70 \pm 0.21$ &
        $86.68 \pm 0.09$
         \\
        
        GCN &
        $45.65 \pm 0.04$ &
        $46.02 \pm 0.26$  &
        $87.42 \pm 0.37$ \\
        
        GAT &
        $45.37 \pm 0.44$ &
        $46.05 \pm 0.51$ &
        $55.80 \pm 0.87$  \\
        
        MixHop &
        $52.16 \pm 0.09$ &
        $51.81 \pm 0.17$  &
        $90.58 \pm 0.16$ \\
        
        LINKX &
        $\bf \color{violet} 61.95 \pm 0.12$ &
        $\bf \color{blue} 56.00 \pm 1.34$ &
        $\bf \color{blue} 90.77 \pm 0.27$ \\
        
        LINK &
        $60.39 \pm 0.07$ &
        $53.97 \pm 0.18$  &
        $73.56 \pm 0.14$ \\
        
        GCNII &
        $37.88 \pm 0.69$ &
        $47.21 \pm 0.28$ &
        $90.24 \pm 0.09$ \\
        
        APPNP &
        $32.19 \pm 0.07$ &
        $38.15 \pm 0.26$ &
        $85.36 \pm 0.62$ \\
        
        GloGNN &
        $\bf \color{blue} 62.09 \pm 0.27$ &
        $\bf \color{violet} 54.68 \pm 0.34$ &
        $\bf \color{violet} 90.66 \pm 0.11$ \\
        
        GPR-GNN &
        $40.19 \pm 0.03$ &
        $45.07 \pm 0.21$  &
        $90.05 \pm 0.31$ \\
        
        ACM-GCN &
        $55.14 \pm 0.16$ &
        $47.37 \pm 0.59$  &
        $80.33 \pm 3.91$ \\ \midrule
        
        \model-GraphSAGE& 
        $\bf \color{red} 69.50 \pm 0.39$&
        $\bf \color{red} 63.30 \pm 1.84$ &
        $\bf \color{red} 90.85 \pm 0.64$\\
        
        \bottomrule
        \vspace{-5mm}
    \end{tabular}
    }
\end{wraptable}
Given the exceptional performance of \model-GraphSAGE on small and medium sized heterophilic graphs, we test the proposed \model~(applied to GraphSAGE, i.e., \model-GraphSAGE) on large-scale datasets. 
To this end, we consider three different experiments based on large graphs from \citet{linkx}, which range from highly heterophilic (homophily level of 0.07) to fairly homophilic (homophily level of 0.61). 
The sizes range from large graphs with $\sim$170K nodes and $\sim$1M edges to a very large graph with $\sim$3M nodes and $\sim$14M edges.

\Tref{tab:hetero_large} shows the results of \model-GraphSAGE together with other standard GNNs, as well as recent state-of-the-art models, i.e., MLP\citep{dlbook}, GCN \citep{gcn}, GAT \citep{gat}, MixHop \citep{mixhop}, LINK(X) \citep{linkx}, GCNII \citep{gcnii}, APPNP \citep{appnp}, GloGNN \citep{glognn}, GPR-GNN \citep{gpr_gnn} and ACM-GCN \citep{acm_gcn}. 
We can see that \model-GraphSAGE significantly outperforms current state-of-the-art (by up to $13\%$) on the two heterophilic graphs (i.e., snap-patents and arXiv-year). 
Moreover, \model-GraphSAGE is on-par with the current state-of-the-art on the homophilic graph dataset genius.

We conclude that the proposed gradient gating method can successfully be scaled up to large graphs, reaching state-of-the-art performance, in particular on heterophilic graph datasets.

\section{Related Work}
\paragraph{Gating.} 
Gating is a key component of our proposed framework. 
The use of gating (i.e., the modulation between $0$ and $1$) of hidden layer outputs has a long pedigree in neural networks and sequence modeling. %
In particular, classical  recurrent neural network (RNN) architectures such as LSTM \citep{lstm} and GRU \citep{gru} rely on gates to modulate information propagation in the RNN. 
Given the connections between RNNs and early versions of GNNs \citep{zhou}, it is not surprising that the idea of gating has been used in designing GNNs \citet{gated_gcn,gated_gcn_2,gaan}. 
However, to the best of our knowledge, the use of local graph-gradients to further modulate gating 
%in \model, 
in order to alleviate the oversmoothing problem is novel, and so is its theoretical analysis. 

\paragraph{Multi-scale methods.}
The multi-rate gating procedure used in \model~is a particular example of \emph{multi-scale} mechanisms. The use of multi-scale neural network architectures  has a long history. An early example is \citet{HinPla}, who proposed a neural network with each connection having a fast changing weight for temporary memory and a slow changing weight for long-term learning. 
The classical convolutional neural networks (CNNs, \citet{lecun1989backpropagation}) can be viewed as multi-scale architectures for processing multiple \emph{spatial} scales in images \citep{Kolter}. Moreover, there is a close connection between our multi-rate mechanism~\eqref{eq:ms_mp_non_oversmoothing} and the use of multiple time scales in recently proposed sequence models such as UnICORNN \citep{unicornn} and long expressive memory (LEM) \citep{lem}. 

\paragraph{Neural differential equations.} Ordinary and partial differential equations (ODEs and PDEs) are playing an increasingly important role in designing, interpreting, and analyzing novel graph machine learning architectures  
\cite{Ave1,Poli2019,Zhu1,Xhonneux2020}. \citet{grand} designed attentional GNNs by discretizing parabolic diffusion-type PDEs. \citet{di2022graph} interpreted GCNs as gradient flows minimizing a generalized version of the Dirichlet energy.  \citet{blend} applied a non-Euclidean diffusion equation (``Beltrami flow'') yielding a scheme with adaptive spatial derivatives (``graph rewiring''), and \citet{topping2021understanding} studied a discrete geometric PDE similar to Ricci flow to improve information propagation in GNNs.  \citet{pde-gcn} proposed a GNN framework arising from a mixture of parabolic (diffusion) and hyperbolic (wave) PDEs on graphs with convolutional coupling operators, which describe dissipative wave propagation. 
Finally, \citet{graphcon} used  systems of nonlinear oscillators coupled through the associated graph structure to rigorously overcome the oversmoothing problem. In line with these works, one contribution of our paper is  a continuous version of \model~\eqref{eq:ggode}, which we used  for a rigorous analysis of the oversmoothing problem. 
Understanding whether this system of ODEs has an interpretation as a  known physical model is a topic for future research.

%\vspace{-3mm}
\section{Discussion}
%\vspace{-3mm}

We have proposed a novel framework, termed \model, for efficient learning on graphs. \model~builds on standard MPNNs, but seeks to overcome their limitations. In particular, we focus on the fact that for standard MPNNs such as GCN or GAT, each node (in every hidden layer) is updated at the same \emph{rate}. This might inhibit efficient learning of tasks where different node features would need to be updated at different rates. Hence, we equip a standard MPNN with \emph{gates} that amount to a  multi-rate modulation for the hidden layer output in \eqref{eq:ms_mp_non_oversmoothing}. This enables multiple rates (or scales) of flow of information across a graph. Moreover, we leverage local (graph) gradients to further constrain the gates. This is done to alleviate oversmoothing where node features become indistinguishable as the number of layers is increased. 

By combining these ingredients, we present a very flexible framework (dubbed \model) for graph machine learning wherein \emph{any} existing MPNN hidden layer can be employed as the coupling function and the multi-rate gradient gating mechanism can be built on top of it. 
Moreover, we also show that \model~corresponds to a time-discretization of a system of ODEs \eqref{eq:ggode}. 
By studying the (in)-stability of the corresponding zero-Dirichlet energy steady states we rigorously prove that gradient gating can mitigate the oversmoothing problem, paving the way for the use of very deep GNNs within the \model~framework. 
In contrast, the lack of gradient gating is shown to lead to oversmoothing. 

We also present an extensive empirical evaluation to illustrate different aspects of the proposed \model~framework. 
Starting with synthetic, small-scale experiments, we demonstrate that 
i) \model~can prevent oversmoothing by keeping the Dirichlet energy constant, even for a very large number of hidden layers, 
ii) this feature allows us to deploy very deep architectures and to observe that the accuracy of classification tasks can \emph{increase} with increasing number of hidden layers, 
iii) the multi-rate mechanism significantly improves performance on node regression tasks when the node features are distributed over a range of scales, and iv) 
\model~is very accurate at classification on \emph{heterophilic} datasets, witnessing an increasing gain in performance with increasing heterophily.  

This last feature was more extensively investigated, and we observed that \model~can significantly outperform baselines as well as recently proposed methods on both benchmark medium-scale and large-scale heterophilic datasets, achieving state-of-the-art performance. 
Thus, by a combination of theory and experiments, we demonstrate that the \model-framework is a promising approach for learning on graphs. 

%\vspace{-3mm}

\paragraph{Future work.}
As future work, we would like to better understand the continuous limit of \model, i.e., the ODEs \eqref{eq:ggode}, especially in the zero spatial-resolution limit and investigate if the resulting continuous equations have interesting geometric and analytical properties. 
Moreover, we would like to use \model~for solving scientific problems, such as in computational chemistry or the numerical solutions of PDEs. 
Finally, the promising results for \model~on large-scale graphs encourage us to use it for even larger industrial-scale applications. 

\section*{Acknowledgements} 
The research of TKR and SM was performed under a project that has received funding from the European Research Council (ERC) under the European Union’s Horizon 2020 research and innovation programme (Grant Agreement No. 770880). 
MWM would like to acknowledge the IARPA (contract W911NF20C0035), NSF, and ONR for providing partial support of this work.
MB is supported in part by ERC Grant No. 724228 (LEMAN). 
The authors thank Emmanuel de Bézenac (ETH) for his constructive suggestions.

\bibliography{refs}
\bibliographystyle{iclr2023_conference}

\newpage
\appendix
\begin{center}
{\large \bf Supplementary Material for:}\\
Gradient Gating for Deep Multi-Rate Learning on Graphs
\end{center}

\section{Additional experiments}

In this section, we describe additional empirical results to complement those in the main text.

\paragraph{On the multi-rate effect of \model.} 
Here, we analyze the performance of \model~on the multi-scale node-level regression task of the main text. 
As we see in the main text, \model~applied to GCN or GAT outperforms their plain counterparts (GCN and GAT) on the multi-scale node-level regression task by more than 50\% on Chameleon and more than 100\% on Squirrel. 
The question therefore arises whether this better performance can be explained by the multi-rate nature of gradient~gating.

To empirically analyse this, we begin by adding a control parameter $\alpha$ to  \model~\eqref{eq:ms_mp_non_oversmoothing} as follows,
\begin{equation*}
    \bX^n =  (1 - \left(\boldsymbol{\tau}^n\right)^{\alpha})\odot \bX^{n-1} + \left(\boldsymbol{\tau}^n\right)^{\alpha} \odot \sigma(\bF_\theta(\bX^{n-1},\cG)).
\end{equation*}

Clearly, setting $\alpha=1$ recovers the original gradient gating message-passing update,
\begin{equation*}
    \bX^n =  (1 - \boldsymbol{\tau}^n)\odot \bX^{n-1} + \boldsymbol{\tau}^n\odot \sigma(\bF_\theta(\bX^{n-1},\cG)),
\end{equation*}
while setting $\alpha=0$ disables any explicit multi-rate behavior and a plain message-passing scheme is recovered,
\begin{equation*}
    \bX^n = \sigma(\bF_\theta(\bX^{n-1},\cG)).
\end{equation*}
Note that by continuously changing $\alpha$ from 0 to 1 controls the level of multi-rate behavior in the proposed gradient gating method.

\begin{figure}[ht!]
\centering
\begin{minipage}[t]{.475\textwidth}
\includegraphics[width=1.\textwidth]{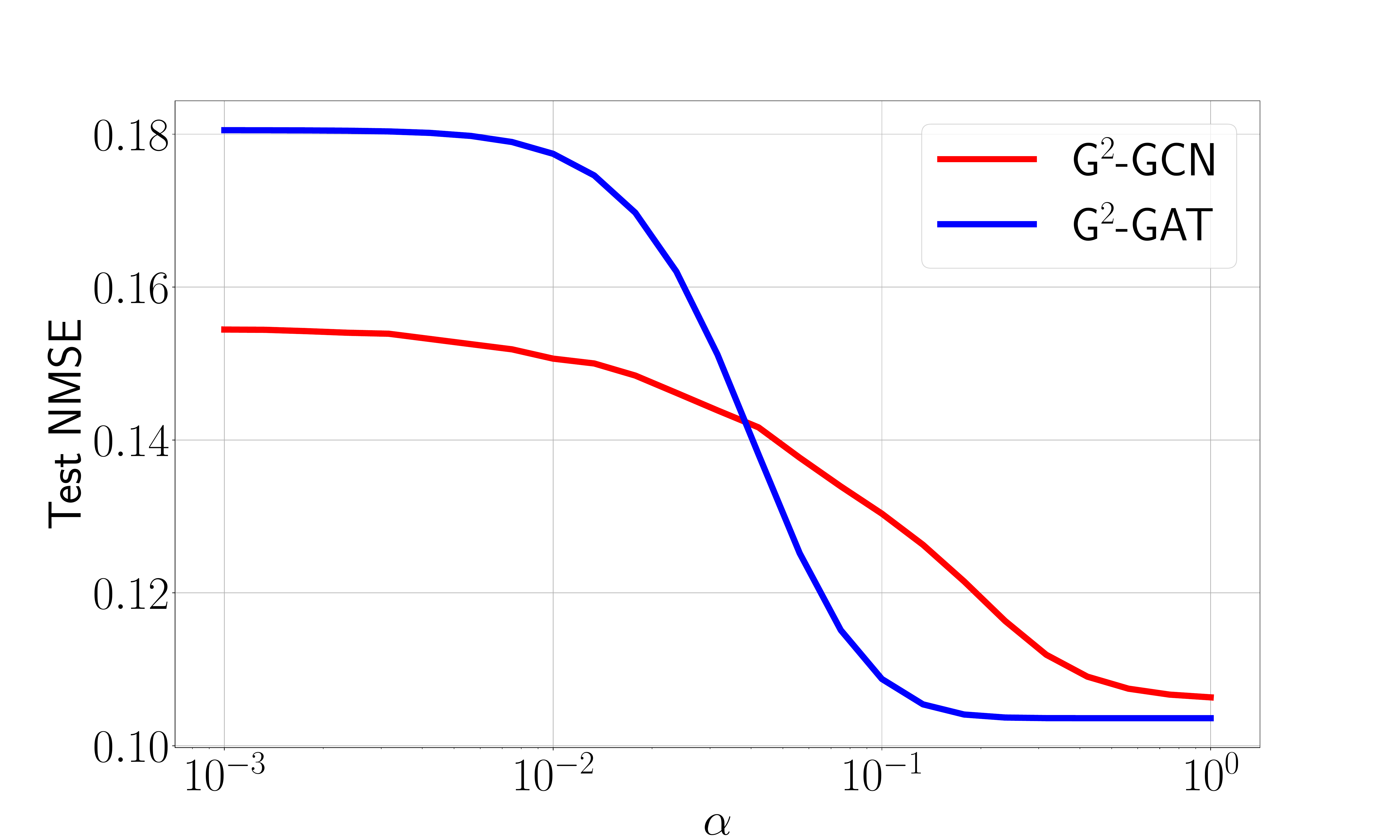}
\caption{Test NMSE on the multi-scale chameleon node-level regression task of \model-GCN and \model-GAT for continuously decreasing level of multi-rate behavior.}
\label{fig:ms_effect_wiki}
\end{minipage}%
\hfill
\begin{minipage}[t]{.475\textwidth}
\includegraphics[width=1.\textwidth]{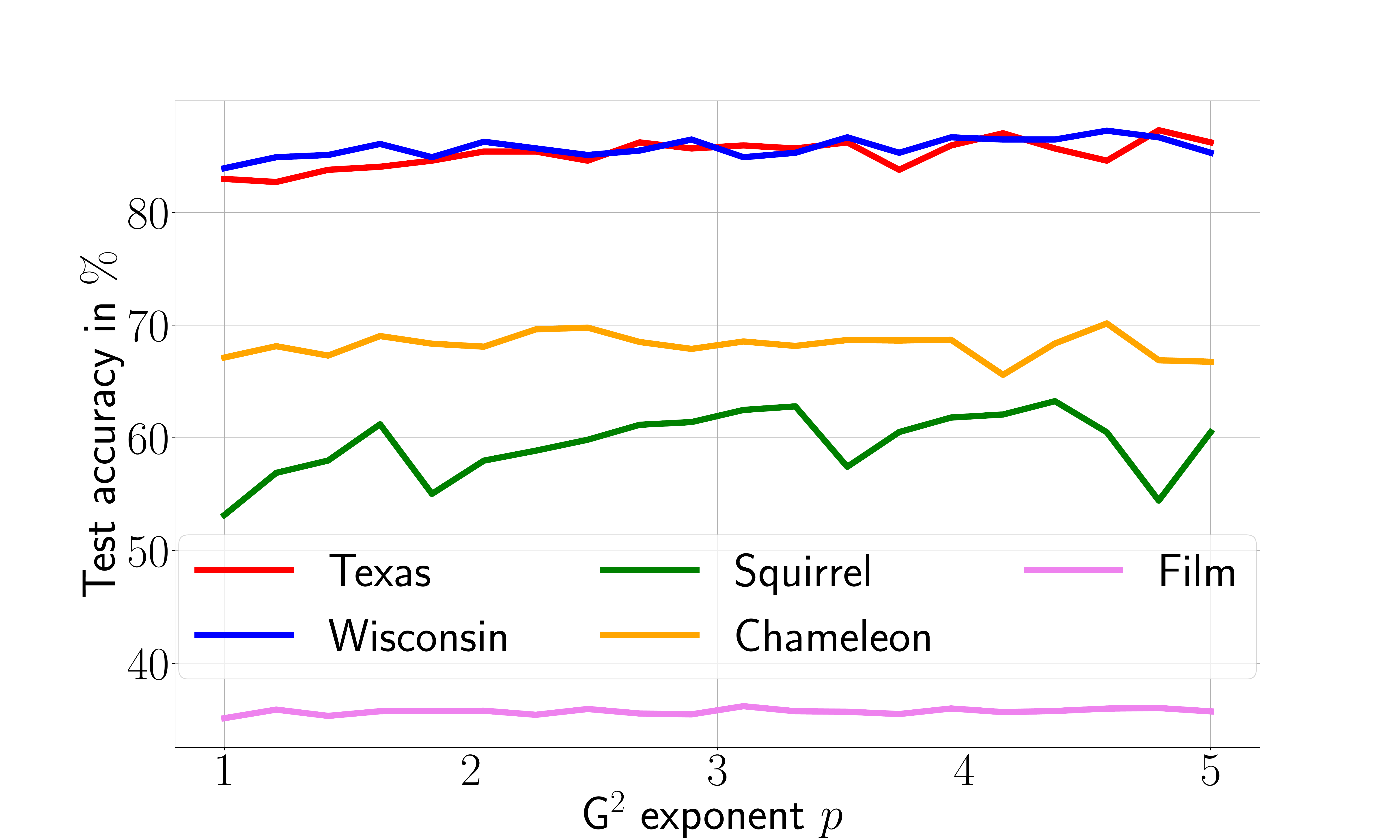}
\caption{Test accuracies of \model-GraphSAGE on Texas, Squirrel, Film, Wisconsin and Chameleon graph datasets for varying values of $p$ in \eqref{eq:ms_mp_non_oversmoothing}.}
\label{fig:grad_exp_sensitivity}
\end{minipage}%
\end{figure}

In \fref{fig:ms_effect_wiki} we plot the test NMSE of the best performing \model-GCN and \model-GAT on the Chameleon multi-scale node-level regression task for increasing values of $\alpha\in[10^{-3},1]$ in log-scale. We can see that the test NMSE monotonically decreases (lower error means better performance) for both \model-GCN and \model-GAT for increasing values of $\alpha$, i.e., increasing level of multi-rate behavior.
We can conclude that the multi-rate  behavior of \model~is instrumental in successfully learning multi-scale regression tasks.

\paragraph{On the sensitivity of performance of \model~to the hyperparameter $p$.}
The proposed gradient gating model implicitly depends on the hyperparameter $p$, which defines the multiple rates $\boldsymbol{\tau}$, i.e.,
\begin{equation*}
\begin{aligned}
    \hat{\boldsymbol{\tau}}^n &= \hat{\sigma}(\hat{\bF}_\theta(\bX^{n-1},\cG)),\\
    \boldsymbol{\tau}^n_{ik} &= 
    \tanh\left(\sum_{j\in\cN_i}|\hat{{\tau}}^n_{ik} - \hat{{\tau}}^n_{jk}|^p\right).
\end{aligned}
\end{equation*}
While any value $p>0$ can be used in practice, a standard hyperparameter tuning procedure (see \ref{train_details} for the training details) on $p$ has been applied in every experiment included in this paper. 
Thus, it is natural to ask how sensitive the performance of \model~is with respect to different values of the hyperparameter $p$.

To answer this question, we trained different \model-GraphSAGE models on a variety of different graph datasets (i.e., Texas, Squirrel, Film, Wisconsin and Chameleon) for different values of $p\in[1,5]$. \fref{fig:grad_exp_sensitivity} shows the resulting performance of \model-GraphSAGE. We can see that different values of $p$ do not significantly change the performance of the model. However, including the hyperparameter $p$ to the hyperparameter fine-tuning procedure will further improve the overall performance of \model.

\begin{figure}[h]
\centering
\begin{minipage}{.475\textwidth}
\includegraphics[width=1.\textwidth]{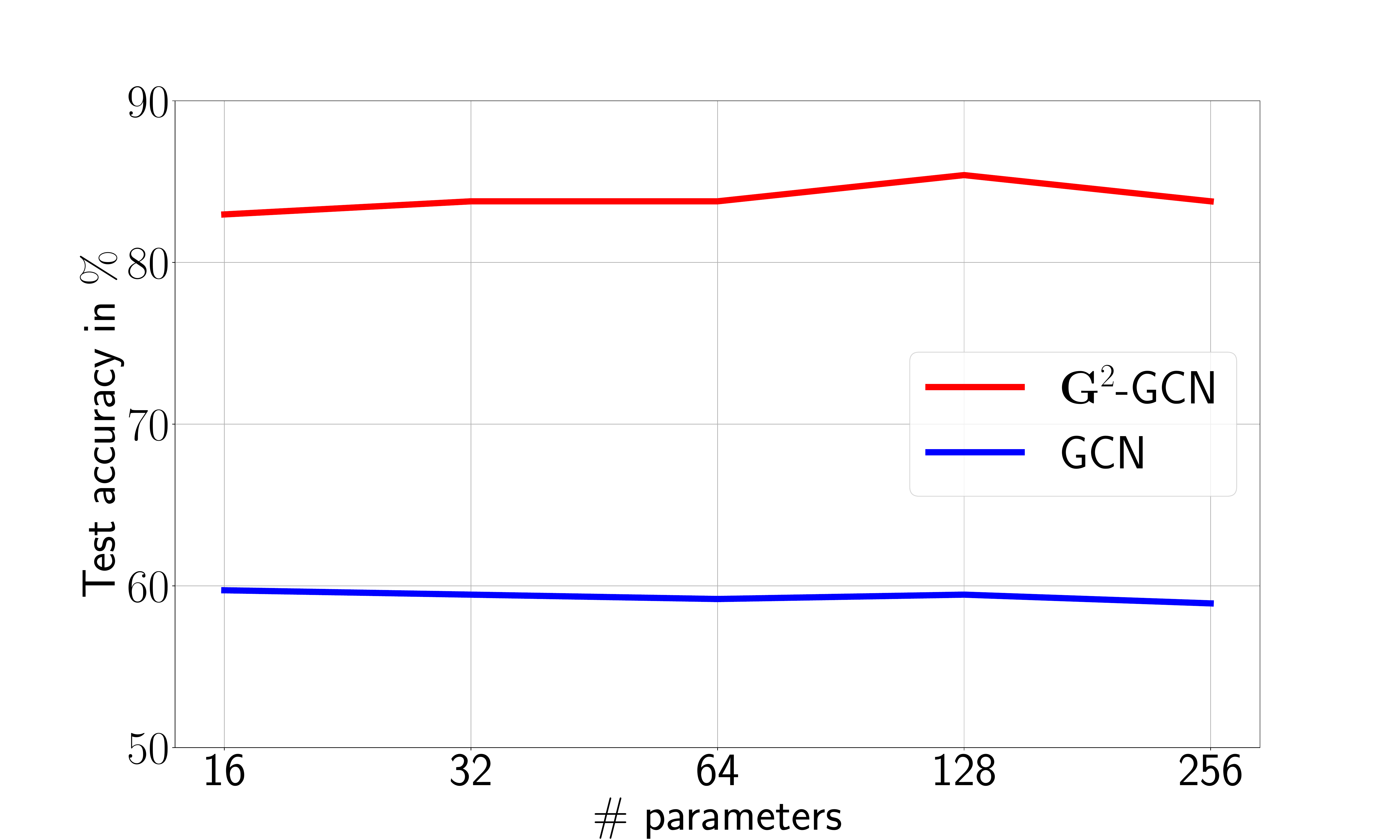}
\caption{Test accuracies of plain GCN and \model-GCN on Texas for varying number of total parameters in the GNN.}
\label{fig:fixed_params}
\end{minipage}
\end{figure}

\paragraph{On the sensitivity of performance of \model~to the number of parameters.}
All results of \model~provided in the main paper are obtained using standard hyperparameter tuning (i.e., random search). Those hyperparameters include the number of hidden channels for each hidden node of the graph, which directly correlates with the total number of parameters used in \model. It is thus natural to ask how \model~performs compared to its plain counter-version (e.g. \model-GCN to GCN) for the exact same number of total parameters of the underlying model. To this end, \fref{fig:fixed_params} shows the test accuracies of \model-GCN and GCN for increasing number of total parameters in its corresponding model. We can see that first, using more parameters has only a slight effect on the overall performance of both models. Second, and most importantly, \model-GCN constantly reaches significantly higher test accuracies for the exact same number of total parameters. We can thus rule out that the better performance of \model~compared to its plain counter-versions is explained by the usage of more parameters.

\paragraph{Ablation of $\hat{\bF}_\theta$ in \model.}
In its most general form \model~\eqref{eq:ms_mp_non_oversmoothing} uses an additional GNN $\hat{\bF}_\theta$ to construct the multiple rates $\boldsymbol{\tau}^n$. Is this additional GNN needed ? To answer this question, \Tref{tab:additional_gnn} shows in which of the provided experiments (using \model-GraphSAGE) we actually used an additional GNN $\hat{\bF}_\theta$ (as suggested by our hyperparameter tuning protocol). We can see that on small-scale experiments having an additional GNN is not needed. However, on the considered medium and large-scale graph datasets it is beneficial to use it. Motivated by this, \Tref{tab:w_vs_wo_add_gnn} shows the results for \model-GraphSAGE on the three medium-scale graph datasets (Film, Squirrel and Chameleon) without using an additional GNN in \eqref{eq:ms_mp_non_oversmoothing} (i.e., $\bF_\theta = \hat{\bF}_\theta$) as well as with using an additional GNN (i.e., the results provided in the main paper). We can see that while \model-GraphSAGE without an additional GNN (i.e., w/o $\hat{\bF}_\theta$) yields competitive results, using an additional GNN is needed in order to obtain state-of-the-art results on these three datasets. 
\begin{table*}[h!]
    \centering
    \caption{Usage of $\hat{\bF}_\theta$ in \model~\eqref{eq:ms_mp_non_oversmoothing} for each result with \model-GraphSAGE presented in the main paper (YES indicates the usage of $\hat{\bF}_\theta$, while NO indicates that no additional GNN is used to construct the multiple rates, i.e., $\bF_\theta = \hat{\bF}_\theta$)
    }
    \label{tab:additional_gnn}
    \resizebox{\textwidth}{!}{%
    \begin{tabular}{ccccccccc}
    \toprule 
         \textbf{Texas} &  
         \textbf{Wisconsin} & 
         \textbf{Film} &
         \textbf{Squirrel} &
         \textbf{Chameleon} &
         \textbf{Cornell} &
         \textbf{snap-patents} &
         \textbf{arXiv-year} &
         \textbf{genius} \\
        
         \midrule
         
         NO & NO & YES & YES & YES & NO & YES & YES & YES
         \\
        
        \bottomrule
    \end{tabular}
    }
\end{table*}

\begin{table*}[h!]
    \centering
    \caption{Test accuracies of \model-GraphSAGE with and without additional GNN (i.e., w/ $\hat{\bF}_\theta$ and w/o $\hat{\bF}_\theta$ in \eqref{eq:ms_mp_non_oversmoothing}) on Film, Squirrel and Chameleon graph dataset.  
    }
    \label{tab:w_vs_wo_add_gnn}
    %\resizebox{\textwidth}{!}{%
    \begin{tabular}{l ccc}
    \toprule 
         &
         \textbf{Film} &
         \textbf{Squirrel} &
         \textbf{Chameleon} \\
         \midrule
         
         \model-GraphSAGE w/ $\hat{\bF}_\theta$ &
         $37.14 \pm 1.01$ &
         $64.26 \pm 2.38$ &
         $71.40 \pm 2.38$ 
         \\
        
        \model-GraphSAGE w/o $\hat{\bF}_\theta$ &
        $36.83 \pm 1.26$&
        $55.78 \pm 1.61$ &
        $65.04 \pm 2.27$
        
         \\
         
        \bottomrule
    \end{tabular}
    %}
\end{table*}

\paragraph{Ablation of multi-rate channels in \model.}
The corner stone of the proposed \model~is the multi-rate matrix $\boldsymbol{\tau}^n$ in \eqref{eq:ms_mp_non_oversmoothing}, which automatically solves the oversmoothing issue for any given GNN (Proposition \ref{prop:3}). This multi-rate matrix learns different rates for every node but also for every channel of every node. It is thus natural to ask if the multi-rate property for the channels is necessary, or if having multiple rates for the different nodes is sufficient, i.e., having a \textbf{multi-rate vector} $\boldsymbol{\tau}^n \in \mathbb{R}^v$. A direct construction of such multi-rate vector (derived from our proposed \model) is:

\begin{equation}
\begin{aligned}
\label{eq:single_scale_channels}
    \hat{\boldsymbol{\tau}}^n &= {\sigma}(\hat{\bF}_\theta(\bX^{n-1},\cG)),\\
    \boldsymbol{\tau}^n_{i} &=
    \tanh\left(\sum_{j\in\cN_i}\|\hat{\boldsymbol{\tau}}^n_{j} - \hat{\boldsymbol{\tau}}^n_{i}\|_p\right), \\
    \bX^n &=  (1 - \boldsymbol{\tau}^n)\odot \bX^{n-1} + \boldsymbol{\tau}^n \odot \sigma(\bF_\theta(\bX^{n-1},\cG)).
\end{aligned}
\end{equation}

Note that the only difference to our proposed \model~is in the second equation of \eqref{eq:single_scale_channels}, where we sum over the node-wise $p$-norms of the differences of adjacent nodes. This way, we compute a single scalar $\boldsymbol{\tau}^n_i$ for every node $i \in \cV$.

\Tref{tab:mr_channel_abl} shows the results of our proposed \model-GraphSAGE as well as the single-rate channels ablation of \model~(eq. \eqref{eq:single_scale_channels}) on the Film, Squirrel and Chameleon graph datasets. As a baseline, we also include the results of a plain GraphSAGE. We can see that while \model~with single-scale channels outperforms the base GraphSAGE model, our proposed \model~with multi-rate channels vastly outperforms the single-rate channels version of \model.

\begin{table*}[h!]
    \centering
    \caption{Test accuracies of plain GraphSAGE, \model-GraphSAGE with multi-rate channels for each node (i.e., standard \model~\eqref{eq:ms_mp_non_oversmoothing}) as well as with only a single rate for every channel on Film, Squirrel and Chameleon.}
    \label{tab:mr_channel_abl}
    %\resizebox{\textwidth}{!}{%
    \begin{tabular}{l ccc}
    \toprule 
         &
         \textbf{Film} &
         \textbf{Squirrel} &
         \textbf{Chameleon} \\
         \midrule
         
         GraphSAGE &
         $34.23 \pm 0.99$ & 
         $41.61 \pm 0.74$ & 
         $58.73 \pm 1.68$ 
         \\
         
         \model-GraphSAGE w/ multi-rate channels \model~\eqref{eq:ms_mp_non_oversmoothing}&
         $37.14 \pm 1.01$ &
         $64.26 \pm 2.38$ &
         $71.40 \pm 2.38$ \\
        
         \model-GraphSAGE w/ single-rate channels \model~\eqref{eq:single_scale_channels} &
         $36.67 \pm 0.56$ &
         $44.03 \pm 1.01$ &
         $60.29 \pm 3.45$
         \\
         
        \bottomrule
    \end{tabular}
    %}
\end{table*}

\begin{figure}[ht!]
\centering
\begin{minipage}[t]{.5\textwidth}
\includegraphics[width=1.\textwidth]{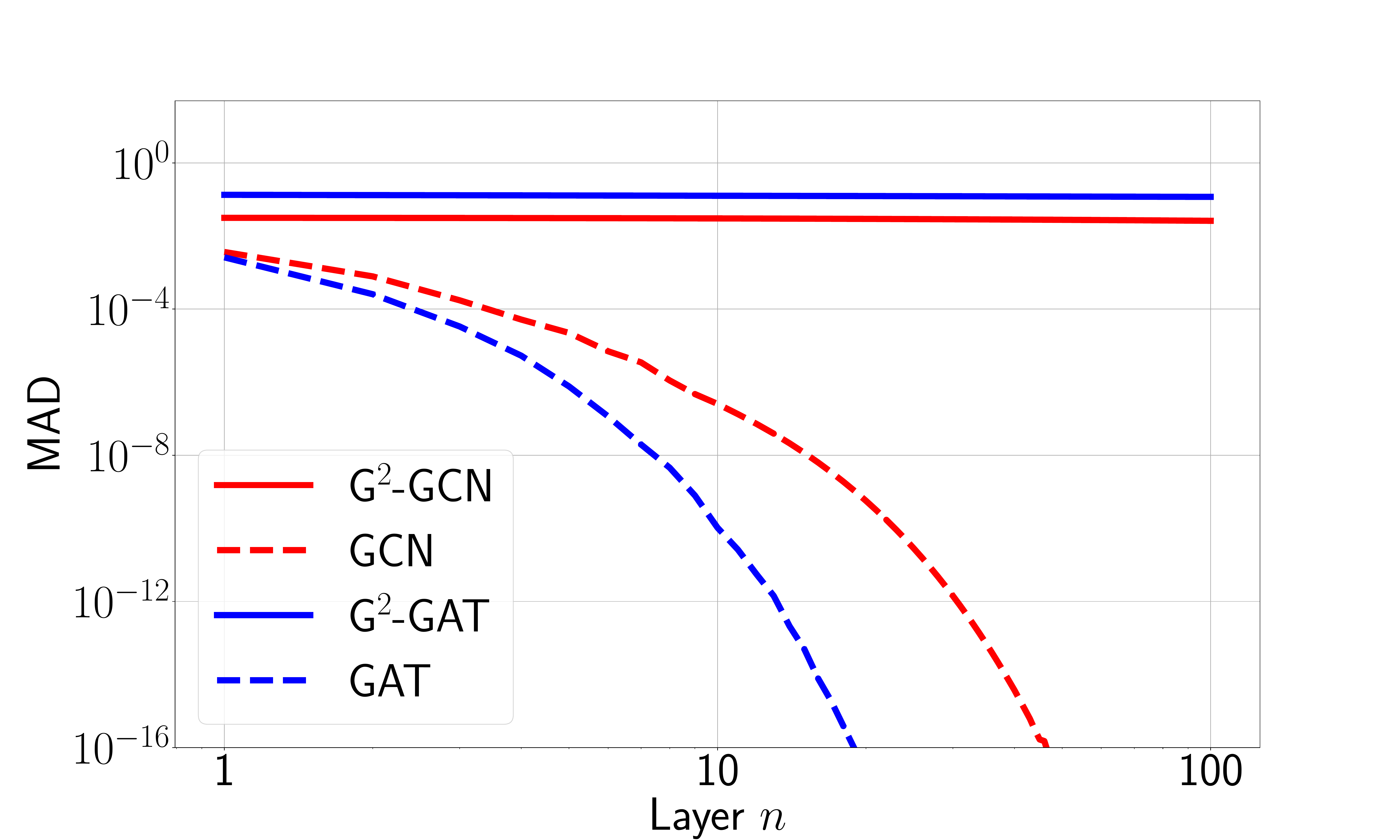}
\caption{Mean average distance (MAD) of layer-wise
node features $\bX^n$ propagated through a GAT,
GCN and their gradient gated versions (\model-GAT, \model-GCN).}
\label{fig:mad}
\end{minipage}
\end{figure}

\paragraph{Alternative measures of oversmoothing.}
The proof of Proposition \ref{prop:2} and Proposition \ref{prop:3} as well as the first experiment in the main paper is based on the definition of oversmoothing using the Dirichlet energy, which was proposed in \citet{graphcon}. However, there exist alternative measures to describe the oversmoothing phenomenon in deep GNNs. One such measure is the mean average distance (MAD), which was proposed in \citet{mad}. In order to check if our proposed \model~mitigates oversmoothing defined through the MAD measure we repeat the first experiment in the main paper and plot the MAD instead of the Dirichlet energy for increasing number of layers in \fref{fig:mad}. We can see that while the MAD of a plain GCN and GAT converges exponentially with increasing number of layers, it remains constant for \model-GCN and \model-GAT. We can thus conclude that \model~mitigates oversmoothing defined through the MAD measure.

\section{Training details}
\label{train_details}
All small and medium-scale experiments have been run on NVIDIA GeForce RTX 2080 Ti, GeForce RTX 3090, TITAN RTX and Quadro RTX 6000 GPUs. The large-scale experiments have been run on Nvidia Tesla A100 (40GiB) GPUs.

All hyperparameters were tuned using random search. \Tref{tab:hp_ranges} shows the ranges of each hyperparameter as well as the random distribution used to randomly sample from it. Moreover,
\Tref{tab:hyps} shows the rounded hyperparameter $p$ in \model~\eqref{eq:ms_mp_non_oversmoothing} of each best performing network. 

\begin{table*}[h!]
    \centering
    \caption{Hyperparameter ranges.}
    \label{tab:hp_ranges}
    %\resizebox{\textwidth}{!}{%
    \begin{tabular}{l|ll}
    \toprule 
         &
         \textbf{range} &
         \textbf{rand. distribution} \\
         \midrule
         
         learning rate &
         $[10^{-4},10^{-2}]$ &
         log uniform
         \\
         
         hidden size $m$ &
         $\{32, 64, 128, 256, 512\}$ &
         disc. uniform \\
         
         dropout input &
         $[0,0.9]$ &
         uniform
         \\
         
         dropout output &
         $[0,0.9]$ &
         uniform
         \\
         
         weight decay &
         $[10^{-8},10^{-2}]$ &
         log uniform
         \\
         
         \model-exponent $p$ &
         $[1,5]$ &
         uniform
         \\
         
         Usage of $\hat{\bF}_\theta$ in \eqref{eq:ms_mp_non_oversmoothing} &
         $\{$YES, NO$\}$ &
         disc. uniform
         \\
        \bottomrule
    \end{tabular}
    %}
\end{table*}

\begin{table*}[h!]
    \centering
    \caption{Rounded hyperparameter $p$ in \model~of each best performing network.
    }
    \label{tab:hyps}
    \resizebox{\textwidth}{!}{%
    \begin{tabular}{l ccccccccc}
    \toprule 
         &
         \textbf{Texas} &  
         \textbf{Wisconsin} & 
         \textbf{Film} &
         \textbf{Squirrel} &
         \textbf{Chameleon} &
         \textbf{Cornell} &
         \textbf{snap-patents} &
         \textbf{arXiv-year} &
         \textbf{genius} \\
        
         \midrule
        
        \model-GAT &
        3.06 &
        1.68 &
        1.23 &
        3.48 &
        3.54 &
        3.54 &
        / &
        / &
        / \\
        
        \model-GCN &
        3.93 &
        2.92 &
        3.79 &
        1.99 &
        1.08 &
        3.87 &
        / &
        / &
        / 
        \\
        
        \model-GraphSAGE &
        4.47 &
        1.14 &
        2.89 &
        3.04 &
        2.00 &
        3.27 &
        1.60 &
        3.40 &
        4.40
        \\
        
        \bottomrule
    \end{tabular}
    }
\end{table*}

\section{Mathematical Details}
In this section, we provide proofs for Propositions \ref{prop:3} and \ref{prop:4} in the main text. We start with the following technical result which is necessary in the subsequent proofs. 

\paragraph{A Poincare Inequality on Connected Graphs.} Poincare inequalities for functions \citep{Evans} bound function values in terms of their gradients. Similar bounds on node values in terms of graph-gradients can be derived and a particular instance is given below,
\begin{proposition}
\label{prop:piog}
Let $\cG = (\cV,\cE)$ be a connected graph and the corresponding (scalar) node features are denoted by $\bY_i \in \R$, for all $i \in \cV$. Let $\bY_1 = 0$. Then, the following bound holds, 
\begin{equation}
    \label{eq:piog}
    \sum\limits_{i\in \cV} \bY^2_i \leq \overline{d}\Delta_1\sum\limits_{i\in \cV}\sum\limits_{j\in \cN_i} |\bY_j-\bY_i|^2,
\end{equation}
where $\overline{d} = \max\limits_{i \in \cV} {\rm deg}(i)$ and $\Delta_1$ is the eccentricity of the node $1$.
\end{proposition}
\begin{proof}
Fix a node $i \in \cV$. By assumption, the graph $\cG$ is connected. Hence, there exists a path connecting $i$ and the node $1$. Denote the shortest path as $\cP(i,1)$. This path can be expressed in terms of the nodes $\ell_{i,1}$ with $0 \leq \ell \leq \delta$, where $0_{i,1}=1$ and $\delta_{i,1}=i$. For any $\ell$, we require $\ell_{i,1} \sim (\ell+1)_{i,1} $. Moreover, $\delta_{i,1}$ is the graph distance between the nodes $i$ and $1$ and $\Delta_1 = \max\limits_{i \in \cV} \delta_{i,1}$ is the eccentricity of the node $1$.  

Given the node feature $\bY_i$, we can rewrite it as,
\begin{align*}
    \bY_i = \bY_1 + \sum\limits_{\ell=0}^{\delta-1} \bY_{(\ell+1)_{i,1}} - \bY_{\ell_{i,1}} = \sum\limits_{\ell=0}^{\delta-1} \bY_{(\ell+1)_{i,1}} - \bY_{\ell_{i,1}},
\end{align*}
as by assumption $\bY_1 = 0$. 

Using Cauchy-Schwartz inequality on the previous identity yields, 
\begin{align*}
     \bY_i^2 \leq \Delta_1 \sum\limits_{\ell=0}^{\delta-1} \left(\bY_{(\ell+1)_{i,1}} - \bY_{\ell_{i,1}}\right)^2  .
\end{align*}
Summing the above inequality over $i \in \cV$ and using the fact that $\ell_{i,1} \sim (\ell+1)_{i,1}$, we obtain the desired Poincare inequality \eqref{eq:piog}.

\end{proof}

\subsection{Proof of Proposition \ref{prop:3} of Main Text}
\label{app:prop3pf}
\begin{proof}
By the definition of exponential stability, we consider a small perturbation around the steady state $\bc$ and study whether this perturbation grows or decays in time. To this end, define the perturbation~as,
\begin{equation}
    \label{eq:pf1}
    \hat{\bX}_i = \bX_i - c, \quad 1 \leq i \leq v.
\end{equation}
A tedious but straightforward calculation shows that these perturbations evolve by the following \emph{linearized} system of ODEs,
\begin{equation}
    \label{eq:pf2}
    \begin{aligned}
      \frac{d\hat{\bX}_i(t)}{dt} &= 
    \sum_{j \in \cN_i} \bA_{ij}(c,c) \left(\hat{\bX}_j - \hat{\bX_i} \right), \quad \forall t,  ~ \forall i \in \cV . \\ 
    \end{aligned}
\end{equation}
Multiplying $\hat{\bx}_i$ to both sides of \eqref{eq:pf2} yields,
\begin{align*}
    \hat{\bX}_i\frac{d\hat{\bX}_i(t)}{dt} &= 
    \sum_{j \in \cN_i} \bA_{ij}(c,c) \hat{\bX_i} \left(\hat{\bX}_j - \hat{\bX}_i \right),\\
\Rightarrow \quad     \frac{d\hat{\bX}^2_i(t)}{dt} &=  \sum_{j \in \cN_i} \bA_{ij}(c,c) \left(\hat{\bX}^2_j - \hat{\bX}^2_i \right) - \sum_{j \in \cN_i} \bA_{ij}(c,c)  \left(\hat{\bX}_j - \hat{\bX}_i \right)^2  .
\end{align*}
Summing the above identity over all nodes $i \in \cV$ yields,
\begin{align*}
    \frac{d}{dt}\sum\limits_{i \in \cV} \hat{\bX}^2_i(t) &= \sum_{i\in \cV} \sum_{j \in \cN_i} \bA_{ij}(c,c) \left(\hat{\bX}^2_j - \hat{\bX}^2_i \right) 
    -  \sum_{i \in \cV} \sum_{j \in \cN_i} \bA_{ij}(c,c)  \left(\hat{\bX}_j - \hat{\bX}_i \right)^2 \\
    &= \frac{1}{2} \sum_{i\in \cV} \sum_{j \in \cN_i} \underbrace{\left(\bA_{ij}(c,c)- \bA_{j,i}(c,c)\right)}_{=0~\eqref{eq:assm2}} \left(\hat{\bX}^2_j - \hat{\bX}^2_i \right)
    - \frac{1}{2} \sum_{i\in \cV} \sum_{j \in \cN_i} \underbrace{\left(\bA_{ij}(c,c)+ \bA_{j,i}(c,c)\right)}_{=2\bA_{ij}~\eqref{eq:assm2}} \left(\hat{\bX}_j - \hat{\bX}_i \right)^2, \\
    &= -\sum_{i\in \cV} \sum_{j \in \cN_i} \bA_{ij}(c,c)  \left(\hat{\bX}_j - \hat{\bX}_i \right)^2, \\
    &\leq - \underline{a} \sum_{i\in \cV} \sum_{j \in \cN_i}   \left(\hat{\bX}_j - \hat{\bX}_i \right)^2, \quad ({\rm by}~\eqref{eq:assm2}), \\
    &\leq -\frac{\underline{a}}{\overline{d}\Delta_1} \sum\limits_{i\in \cV} \hat{\bX}^2_i.
\end{align*}
Here, the last inequality comes from applying the Poincare inequality \eqref{eq:piog} for the perturbations $\hat{\bX}$ and from the fact that by assumption $\hat{\bX}_1 = 0$.

Applying Gr\"onwall's inequality yields,
\begin{equation}
    \label{eq:pf3}
    \sum\limits_{i\in \cV} \hat{\bX}^2_i(t) \leq \sum\limits_{i\in \cV} \hat{\bX}^2_i(0)e^{-\frac{\underline{a}}{\overline{d}\Delta_1}  t} .
\end{equation}

Thus, the initial perturbations around the steady state $\bc$ are damped down exponentially fast and the steady state $\bc$ is exponentially stable implying that this architecture will lead to oversmoothing. 
\end{proof}

\subsection{Proof of Proposition \ref{prop:4} of Main Text}
\label{app:prop4pf}
\begin{proof}
As in the proof of Proposition \ref{prop:3}, we consider small perturbations of form \eqref{eq:pf1} of the steady state $\bc$ and investigate how these perturbations evolve in time. Assuming that the initial perturbations are small, i.e., that there exists an $0 < \epsilon << 1$ such that $\max\limits_{i \in \cV} |\hat{\bx}_i(0)| \leq \epsilon$, we perform a straightforward calculation to obtain that the perturbations (for a short time) evolve with the following \emph{quasi-linearized} system of ODEs, 
\begin{equation}
    \label{eq:ggode4}
    \begin{aligned}
      \frac{d\hat{\bX}_i(t)}{dt} &= \obtau_i(t) \sum_{j \in \cN_i} \bA_{ij}(c,c) \left(\hat{\bX}_j  - \hat{\bX}_i\right), ~ \forall i \in \cV, \\ 
    \obtau_{i}(t) &= \sum_{j\in\cN_i}|\hat{\bX}_{j}(t) - \hat{\bX}_{i}(t)|^p, ~ \forall i \in \cV .
    \end{aligned}
\end{equation}
Note that we have used the fact that $\sigma^\prime(x) = 1$ and $\tanh^{\prime}(0) = 1$ in obtaining \eqref{eq:ggode4} from \eqref{eq:ggode1}. 

Next, we multiply $\hat{\bx}_i$ to both sides of \eqref{eq:ggode4} to obtain,
\begin{equation}
\label{eq:pf4}
\begin{aligned}
    \hat{\bX}_i\frac{d\hat{\bX}_i(t)}{dt} &= 
    \sum_{j \in \cN_i} \bA_{ij}(c,c) \obtau_i \hat{\bX_i} \left(\hat{\bX}_j - \hat{\bX_i} \right),\\
\Rightarrow \quad     \frac{d\hat{\bX}^2_i(t)}{dt} &=  \sum_{j \in \cN_i} \bA_{ij}(c,c) \obtau_i \left(\hat{\bX}^2_j - \hat{\bX}^2_i \right) - \sum_{j \in \cN_i} \bA_{ij}(c,c)  \obtau_i \left(\hat{\bX}_j - \hat{\bX}_i \right)^2 
\end{aligned}
\end{equation}
Trivially, 
$$
|\hat{\bX}_j - \hat{\bX}_i|^p \leq \obtau_i, ~ \forall j \in \cN_i, \quad \forall i .
$$
Applying this inequality to the last line of the identity \eqref{eq:pf4}, we obtain,
\begin{align*}
    \frac{d\hat{\bX}^2_i(t)}{dt} &\leq  \sum_{j \in \cN_i} \bA_{ij}(c,c) \obtau_i \left(\hat{\bX}^2_j - \hat{\bX}^2_i \right) - \sum_{j \in \cN_i} \bA_{ij}(c,c)  \left|\hat{\bX}_j - \hat{\bX}_i \right|^{p+2} .
\end{align*}
Summing the above inequality over $i \in \cV$ leads to,
\begin{align*}
    \frac{d}{dt}\sum\limits_{i\in \cV} \hat{\bX}^2_i(t) &\leq  \sum_{i \in \cV} \sum_{j \in \cN_i} \bA_{ij}(c,c) \obtau_i \left(\hat{\bX}^2_j - \hat{\bX}^2_i \right) - \sum_{i\in \cV}\sum_{j \in \cN_i} \bA_{ij}(c,c)  \left|\hat{\bX}_j - \hat{\bX}_i \right|^{p+2} \\
    &\leq \frac{1}{2}  \sum_{i \in \cV} \sum_{j \in \cN_i} \bA_{ij}(c,c) \left(\obtau_i-\obtau_j\right) \left(\hat{\bX}^2_j - \hat{\bX}^2_i \right) \quad (\bA_{ij} = \bA_{j,i}) \\
    &- \underline{a}  \sum_{i \in \cV} \sum_{j \in \cN_i}  \left|\hat{\bX}_j - \hat{\bX}_i \right|^{p+2} \quad ({\rm from}~\eqref{eq:assm2}) .
\end{align*}
Therefore, we have the following inequality,
\begin{equation}
    \label{eq:pf5}
    \begin{aligned}
     \frac{d}{dt}\sum\limits_{i\in \cV} \hat{\bX}^2_i(t) &\leq T_1 - T_2, \\
     T_1 &= \frac{1}{2}  \sum_{i \in \cV} \sum_{j \in \cN_i} \bA_{ij}(c,c) \left(\obtau_i-\obtau_j\right) \left(\hat{\bX}^2_j - \hat{\bX}^2_i \right) \\
     T_2 &= \underline{a}  \sum_{i \in \cV} \sum_{j \in \cN_i}  \left|\hat{\bX}_j - \hat{\bX}_i \right|^{p+2} .
\end{aligned}
\end{equation}
We analyze the differential inequality \eqref{eq:pf5} by starting with the term $T_1$ in \eqref{eq:pf5}. We observe that this term does not have a definite sign and can be either positive or negative. However, we can upper bound this term in the following manner. Given that the right-hand side of the ODE system \eqref{eq:ggode4} is Lipschitz continuous, the well-known Cauchy-Lipschitz theorem states that the solutions $\hat{\bx}$ depend continuously on the initial data. Given that $\max\limits_{i \in \cV} |\hat{\bX}_i(0)| \leq \epsilon << 1$ and the bounds on the hidden states \eqref{eq:mp}, there exists a time $\overline{t} > 0$ such that 
$$\max\limits_{i \in \cV} |\hat{\bX}_i(t)| \leq 1, \forall t \in [0,\overline{t}] .$$

Using the definitions of $\obtau$ and the right stochasticity of the matrix $\bA$, we easily obtain the following bound,
\begin{equation}
    \label{eq:pf6} 
    |T_1| \leq 2^{p+1}\overline{d}^2v,
\end{equation}
where $\overline{d} = \max\limits_{i \in \cV} {\rm deg}(i)$.

On the other hand, the term $T_2$ in \eqref{eq:pf5} is clearly positive. Hence, the solutions of resulting ODE, 
\begin{equation}
    \label{eq:pf7}
    \frac{d}{dt}\sum\limits_{i\in \cV} \hat{\bX}^2_i(t) \leq - T_2,
\end{equation}
will clearly decay in time. The key question is whether or not the decay is \emph{exponentially fast}. We answer this question below.

To this end, we have the following calculation using the H\"older's inequality,
\begin{align*}
     \sum_{i \in \cV} \sum_{j \in \cN_i}  \left|\hat{\bX}_j - \hat{\bX}_i \right|^{2} &\leq \left(\overline{d}v\right)^{\frac{p}{p+2}} \left( \sum_{i \in \cV} \sum_{j \in \cN_i}  \left|\hat{\bX}_j - \hat{\bX}_i \right|^{p+2} \right)^\frac{2}{p+2}, \\
\Rightarrow \quad \frac{1}{\left(\overline{d}v\right)^{\frac{p}{2}}}  \left( \sum_{i \in \cV} \sum_{j \in \cN_i}  \left|\hat{\bX}_j - \hat{\bX}_i \right|^{2} \right)^\frac{p+2}{2} &\leq \sum_{i \in \cV} \sum_{j \in \cN_i}  \left|\hat{\bX}_j - \hat{\bX}_i \right|^{p+2}   .
\end{align*}
Observing that $\hat{\bX}_1 =0$ by assumption, we can applying the Poincare inequality \eqref{eq:piog} in the above inequality to further obtain,
\begin{align*}
    \frac{1}{\overline{d}^{p+1} v^{\frac{p}{2}}\Delta_1^{\frac{p+2}{2}}}  \left( \sum_{i \in \cV} |\hat{\bX}_i|^2  \right)^\frac{p+2}{2} &\leq \sum_{i \in \cV} \sum_{j \in \cN_i}  \left|\hat{\bX}_j - \hat{\bX}_i \right|^{p+2}   .
\end{align*}

Hence, from the definition of $T_2$ \eqref{eq:pf5}, we have,
\begin{equation}
    \label{eq:pf8}
    T_2 \geq \frac{\underline{a}}{\overline{d}^{p+1} v^{\frac{p}{2}}\Delta_1^{\frac{p+2}{2}}}  \left( \sum_{i \in \cV} |\hat{\bX}_i|^2  \right)^\frac{p+2}{2} .
\end{equation}
Therefore, the differential inequality \eqref{eq:pf7} now reduces to,
\begin{equation}
    \label{eq:pf9}
    \frac{d}{dt}\sum\limits_{i\in \cV} \hat{\bX}^2_i(t) \leq - \frac{\underline{a}}{\overline{d}^{p+1} v^{\frac{p}{2}}\Delta_1^{\frac{p+2}{2}}}  \left( \sum_{i \in \cV} |\hat{\bX}_i|^2  \right)^\frac{p+2}{2} .
\end{equation}
The differential inequality \eqref{eq:pf9} can be explicitly solved to obtain, 
\begin{equation}
    \label{eq:pf10}
    \sum\limits_{i\in \cV} \hat{\bX}^2_i(t) \leq 
    \left(2 + p t \frac{\underline{a}}{\overline{d}^{p+1} v^{\frac{p}{2}}\Delta_1^{\frac{p+2}{2}}} \sum\limits_{i\in \cV} \hat{\bX}^2_i(0)^{\frac{p}{2}}\right)^{-\frac{2}{p}}  \sum\limits_{i\in \cV} \hat{\bX}^2_i(0) .
\end{equation}
From \eqref{eq:pf10}, we see that the initial perturbations decay but only \emph{algebraically} at a rate of $t^{-\frac{2}{p}}$ in time. For instance, the decay is only linear in time for $p=2$ and even slower for higher value of $p$.

Combining the analysis of the terms $T_{1,2}$ in the differential inequality \eqref{eq:pf5}, we see that the one of the terms can lead to a growth in the initial perturbations whereas the second term only leads to polynomial decay. Even if the contribution of the term $T_1 \equiv 0$, the decay of initial perturbations is only polynomial. Thus, the steady state $\bc$ is not exponentially stable.
\end{proof}

\begin{remark}
We note that the Proposition \ref{prop:4} assumes a certain structure of the matrix $\bA$ in \eqref{eq:ggode1}. A careful perusal of the proof presented above reveal that this assumptions can be further relaxed. To start with, if the matrix $\bA(c,c)$ is not symmetric, then there will be an additional term in the inequality \eqref{eq:pf5}, which would be proportional to $\bA_{ij} - \bA_{ji}$. This term will be of indefinite sign and can cause further growth in the perturbations of the steady state $c$. In any case, it can only further destabilize the quasi-linearized system. The assumption that the entries of $\bA$ are uniformly positive amounts to assuming positivity of the weights of the underlying GNN layer. This can be replaced by requiring that the corresponding eigenvalues are uniformly positive. If some eigenvalues are negative, this will cause further instability and only strengthen the conclusion of lack of (exponential) stability. Finally, the assumption that one node is not perturbed during the quasi-linearization is required for the Poincare inequality \eqref{eq:piog}. If this is not true, an additional term, of indefinite sign, is added to the inequality \eqref{eq:pf5}. This term can cause further growth of the perturbations and will only add instability to the system. Hence, all the assumptions in Proposition \ref{prop:4} can be relaxed and the conclusion of lack of exponential stability of the zero-Dirichlet energy steady state still holds. 
\end{remark}

\end{document}